\documentclass[runningheads]{llncs}

\usepackage{eccv}

\usepackage{eccvabbrv}

\usepackage{graphicx}
\usepackage{booktabs}
\usepackage{comment}
\usepackage{dsfont}

\usepackage[accsupp]{axessibility}

\usepackage[pagebackref,breaklinks,colorlinks,citecolor=eccvblue]{hyperref}

\usepackage{orcidlink}

\newcommand{\modelname}{SEMAGIC\xspace}
\newcommand{\titlename}{\modelname: Learning Semantically Consistent Deformable 3D Representations from \\ In-the-Wild Images}

\begin{document}

\title{\titlename} 

\titlerunning{\modelname: Semantically Consistent Deformable 3D Representations}

\author{Sky Cen\inst{1} \and
Wufei Ma\inst{1} \and
Guofeng Zhang\inst{1} \and
Alan Yuille\inst{1} \and
Adam Kortylewski\inst{2}}

\authorrunning{S. Cen et al.}

\institute{Johns Hopkins University \and
CISPA Helmholtz Center for Information Security}

\maketitle

\begin{abstract}
Learning deformable 3D object models from single-view in-the-wild images has enabled impressive 3D shape reconstruction without supervision. However, it remains unclear whether these models capture the semantic structure required for downstream tasks. We find that existing deformable reconstruction approaches, despite producing visually plausible geometry, yield unstable correspondences across instances and perform poorly on semantic correspondence benchmarks.
We introduce \modelname, a framework for learning semantically consistent deformable 3D representations from single-view in-the-wild images. Rather than treating reconstruction as the end goal, \modelname uses deformable modeling as a mechanism to discover category-level correspondences. 
Each category is represented by a canonical template mesh and a learned deformation field, functioning similarly to an autoencoder that reconstructs instance geometry from image features, enabling vertices to maintain consistent semantic meaning across instances. Semantic consistency is enforced during training through (i) a feature-level consistency loss aligning semantic features between canonical and deformed meshes, and (ii) vertex-index–conditioned deformation that preserves semantic correspondence across instances.
By explicitly coupling geometric deformation with semantic alignment, \modelname produces representations that maintain stable part correspondences across intra-category variation. 
Experiments demonstrate that \modelname improves semantic correspondence of deformable models by $+14.7$ PCK\texttt{@}0.1 on SPair-71k, establishing deformable models as effective semantic 3D representations.
\end{abstract}

\section{Introduction}
\label{sec:intro}

Deformable 3D reconstruction from in-the-wild images has recently become practical, enabling category-level models that recover detailed geometry and plausible pose from only single-view supervision. Methods in this family learn a canonical template and predict instance-specific deformations, yielding impressive reconstructions across challenging Internet imagery and broad intra-class variation \cite{wu2023magicpony,Pavllo_2023_CVPR, li2024fauna}. At the same time, the success of these approaches is largely measured through visual fidelity and reconstruction metrics, leaving an important question unresolved: do the learned models encode \emph{semantic structure} — \ie, do they induce stable correspondences of object parts across instances — a key requirement for downstream tasks \eg in robotics or augmented reality \cite{min2020spair71k,caron2021dino,oquab2023dinov2}? 
In this work, we show that, despite accurate geometry, existing deformable reconstruction models often exhibit \emph{semantic drift} in their canonical parameterization, where the same vertex index may map to different semantic parts across instances, limiting their utility beyond reconstruction.

Yet, deformable models are, in principle, well suited for learning semantic structure: by mapping diverse object instances into a shared canonical space, they implicitly define dense correspondences across a category. Ideally, each vertex in the canonical template would represent a consistent object part, allowing geometry learned from reconstruction to directly support semantic reasoning. In practice, however, existing approaches optimize primarily for photometric or geometric reconstruction, leaving the canonical parameterization underconstrained with respect to semantic identity. As a result, vertices may drift across semantically unrelated regions during training, preventing stable correspondence from emerging despite accurate geometry. 

\begin{figure}
    \centering
    \includegraphics[width=\linewidth]{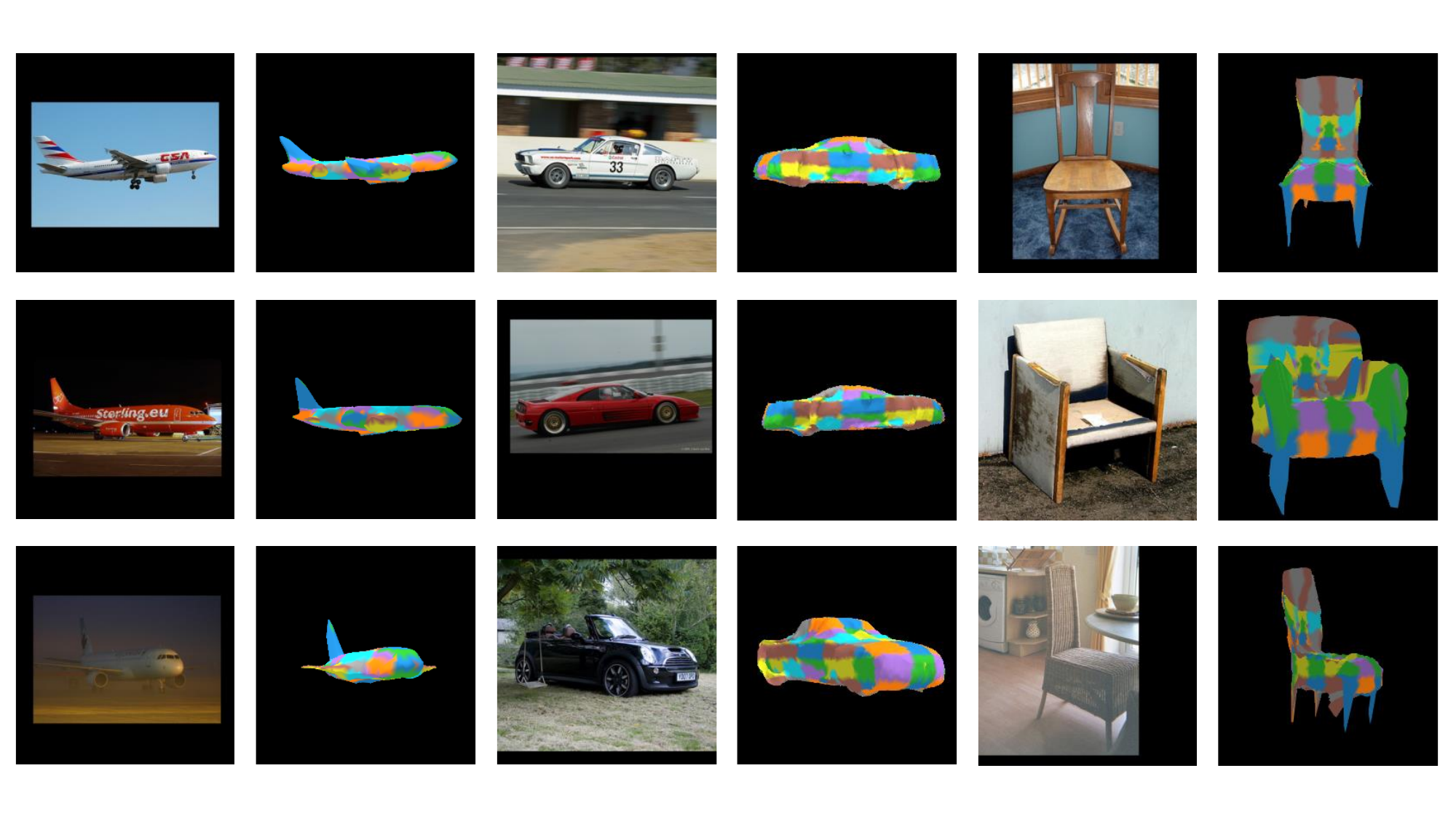}
    \caption{\textbf{Overview of \modelname.} We present \modelname, a method for learning semantically consistent deformable 3D representations from in-the-wild images. Our method learns a category-level canonical mesh and an instance level deformation field that produces dense, semantically consistent correspondences across object instances.}
    \label{fig:teaser}
\end{figure}

In this work, we revisit deformable reconstruction from a semantic perspective and propose \textbf{\modelname}, a framework that treats deformation not merely as a tool for matching geometry, but as a mechanism for discovering category-level semantic correspondences (Figure \ref{fig:teaser}). Our key idea is to explicitly preserve vertex identity across deformations and enforce consistency between canonical and instance representations, transforming the canonical space into a semantically meaningful reference frame shared across object instances.

\modelname learns semantically consistent deformable 3D representations by coupling a canonical geometric prior with semantic constraints during deformation learning. Each object category is represented by a canonical template mesh together with an image-conditioned deformation field that maps the template to individual instances (analogous to an autoencoder that reconstructs 3D geometry from image embeddings). To encourage the canonical space to acquire semantic meaning, \modelname enforces consistency at both the geometric and feature levels: vertex identities are preserved across deformations to maintain persistent correspondence, while semantic image features guide aligned regions of different instances toward similar representations. In addition, we leverage foundation-model predictions for depth and pose as auxiliary supervision to stabilize geometry and resolve ambiguities common in rigid object categories, allowing semantic constraints to operate on reliable reconstructions. Through this combination, deformation becomes a structured alignment process rather than purely geometric fitting, enabling dense correspondences to emerge naturally from the learned representation.
In summary, our contributions are threefold:
\begin{itemize}
    \item We provide an empirical analysis showing that existing deformable 3D reconstruction methods, despite accurate geometry, fail to learn semantically consistent canonical representations, leading to unstable cross-instance correspondences.
    \item We introduce SEMAGIC, a framework that reformulates deformable reconstruction as a semantic learning problem by enforcing persistent vertex identity and feature-level consistency during deformation, enabling the canonical space to serve as a shared semantic reference frame.
    \item We demonstrate that the resulting representations produce substantially improved semantic correspondences across object categories learned from single-view in-the-wild images, establishing deformable models as effective semantic 3D representations rather than purely reconstruction tools.
\end{itemize} 

\section{Related Work}
\label{sec:related}

\textbf{Learning 3D Deformable Models from In-the-Wild Images.} Recovering 3D object structure from in-the-wild images has become an active area of research driven by advances in differentiable rendering and neural implicit representations. Early category-level reconstruction methods demonstrated that meaningful 3D structure can be learned without explicit 3D supervision by leveraging multi-view consistency and object-centric inductive biases \cite{kar2015category, tulsiani2016learning}. For example, Kanazawa \etal \cite{kanazawa2018learning} learns deformable meshes for birds from single images using keypoint supervision, while Goel \etal \cite{goel2020shape} extends previous work to reduce reliance on annotated keypoints.

More recent works remove explicit correspondence supervision and instead learn 3D structure directly from image collections. Methods such as \cite{wu2023magicpony} and \cite{chen2020learning} demonstrate that category-level geometry and camera pose can be jointly optimized from in-the-wild data. The introduction of neural radiance fields \cite{mildenhall2021nerf} and subsequent extensions \cite{yu2021pixelnerf, Pavllo_2023_CVPR}, further enabled reconstruction from sparse or single-view inputs by conditioning implicit 3D representations on image features.

A key challenge in category-level reconstruction is modeling intra-class variation. Objects within the same category can exhibit substantial differences in shape, articulation, and topology. To address this, many approaches adopt deformable representations. For instance, \cite{wu2023magicpony} learns a shared canonical template and predicts per-instance deformations, enabling generalization across diverse object instances. These approaches allows dense correspondences to be defined implicitly through shared coordinates or template vertices. However, most reconstruction methods primarily optimize for photometric, silhouette, or multi-view consistency. While these objectives encourage geometric plausibility, they do not explicitly enforce semantic alignment across instances. As a result, the learned canonical space may not guarantee semantically meaningful correspondences between analogous parts. Our work takes advantage of the dense correspondence that approaches like \cite{wu2023magicpony, Pavllo_2023_CVPR} provide for the primary task of semantic correspondence, rather than 3D reconstruction. Our method introduces effective techniques to enforce semantic consistency during training and retrieve 2D correspondences during evaluation.

\noindent\textbf{Semantic Correspondence.} Semantic correspondence aims to establish dense or sparse correspondences between different instances of an object category \cite{min2020spair71k}. Early approaches formulates this task purely in the 2D image domain, relying on hand-crafted features and geometric regularization \cite{liu2010sift, qiu2014scale, ham2016proposal}. More recent methods leverage deep feature representations and attention mechanisms to improve robustness \cite{liu2020semantic, oquab2023dinov2, hedlin2023unsupervised, zhang2023tale}. In particular, self-supervised models such as DINOv2 \cite{oquab2023dinov2} and Stable Diffusion \cite{hedlin2023unsupervised, zhang2023tale} exhibit strong zero-shot semantic correspondence capabilities, despite not being explicitly trained for this task.

Although these 2D methods have substantially advanced semantic correspondence, they remain limited in scenarios involving ambiguous part appearances~\cite{mariotti2024improving} or large viewpoint variations~\cite{zhang2024telling}, where appearance similarity alone is insufficient for reliable matching. To address these challenges, recent works leverage 3D object structure to facilitate correspondence. For example, Sommer \etal \cite{sommer2025common3d} and Dunkel \etal \cite{dunkel2025yourself} train lightweight adapters to refine off-the-shelf features using 3D object geometry, making them more suitable for semantic correspondence. Similar to our approach, Sommer \etal learn a deformable 3D shape representation that generalizes across rigid object categories. However, their method relies on video collections for training, whereas our model is trained from single-view in-the-wild images. Furthermore, our method learns dense correspondence directly from a canonical 3D representation, enabling more robust alignment under occlusion and large viewpoint variation.

\section{Method}
\label{sec:method}

\begin{figure}[t]
    \centering
    \includegraphics[width=\linewidth]{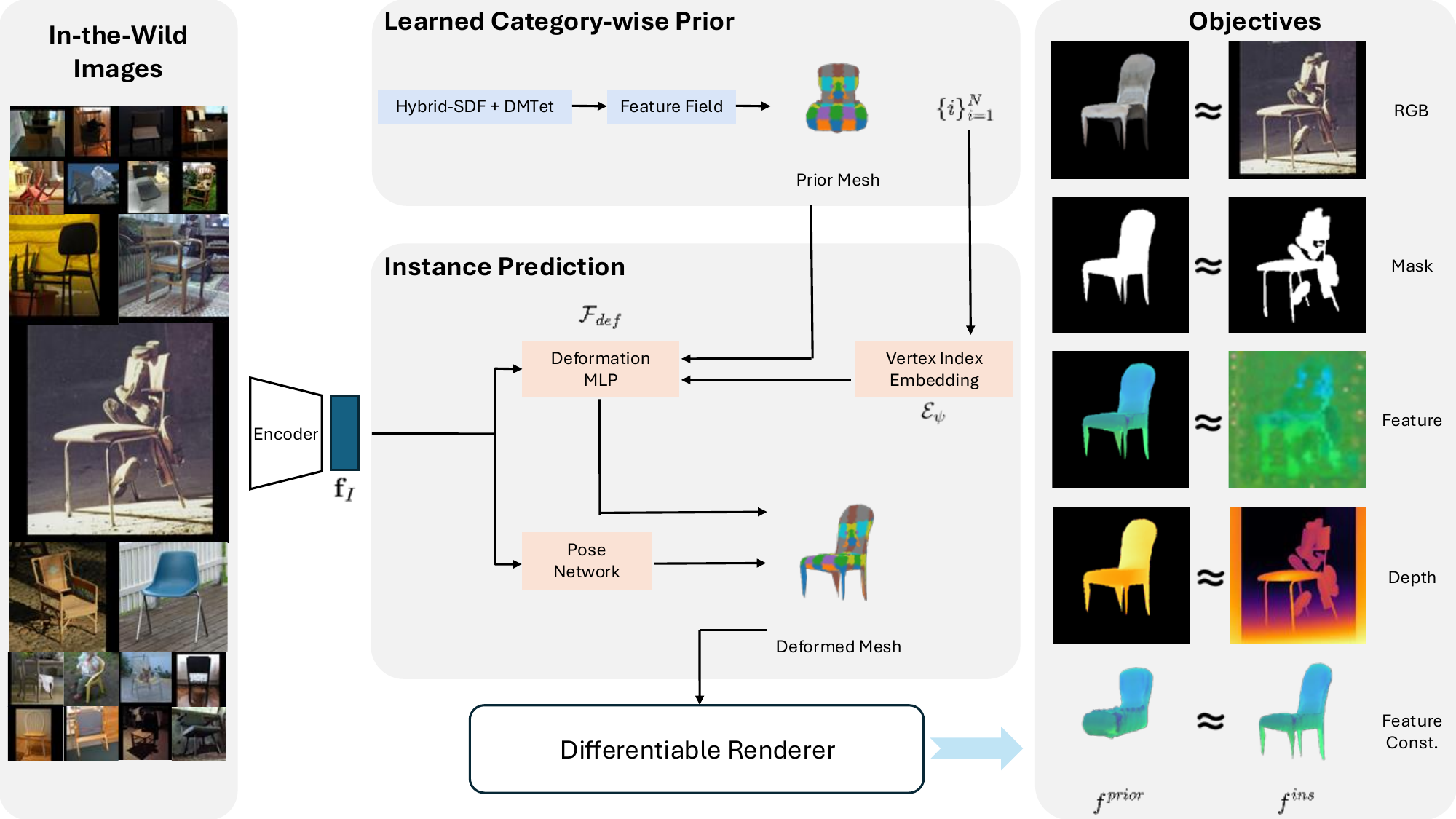}
    \caption{\textbf{Method Overview.} Our method learns a category-level prior by representing shape as a hybrid signed distance field (SDF) optimized via Differentiable Marching Tetrahedra (DMTet). Given an in-the-wild image, the canonical prior is deformed by an MLP $\mathcal{F}_{def}$ conditioned on global image features $\mathbf{f}_I$ extracted from a DINO-ViT encoder, along with vertex index embeddings produced by an embedding network $\mathcal{E}\psi$. A separate pose network predicts the object’s rotation from $\mathbf{f}_I$, which is applied to the deformed mesh before rendering. We also visualize the training objectives: RGB, segmentation mask, feature, depth, and a feature-level consistency loss. The feature-level consistency loss enforces similar feature embeddings for corresponding vertices in the canonical and deformed meshes. By conditioning $\mathcal{F}_{def}$ on vertex identity and incorporating feature-level consistency, \modelname learns a semantically consistent canonical representation.}
    \label{fig:pipeline}
\end{figure}

Deformable 3D reconstruction models show that category-level shape can be recovered from in-the-wild images without explicit 3D supervision. By learning a canonical template mesh and predicting instance-specific deformations, these methods capture impressive geometric variability across diverse object instances. However, existing approaches are optimized primarily for photometric reconstruction and do not explicitly constrain the learned representation to preserve semantic structure. As a result, vertices in the canonical space often drift semantically across different instances. More specifically, the same vertex index may correspond to different object parts depending on the deformation, leading to unstable correspondences and poor performance on semantic tasks.

\modelname rethinks deformable modeling from a semantic perspective. Rather than treating 3D object reconstruction as the final objective, we use deformable reconstruction as a mechanism for discovering dense semantic correspondences across instances.

In this section, we present \modelname, a framework for learning semantically consistent deformable 3D representations. We begin by introducing the method of representing 3D objects in Section~\ref{sec:3.1}. Building on this, in Section~\ref{sec:3.2} we formulate the techniques used to enforce semantic consistency in our \modelname.  Then, in Section~\ref{sec:3.3}, we describe how our \modelname extracts semantic correspondences between two different instances using our learned geometry. Lastly, we discuss the challenges of learning common rigid objects and how our \modelname addresses these challenges with the use of foundational models as supervision in Section~\ref{sec:3.4}. Figure~\ref{fig:pipeline} provides an overview of our method.

\subsection{3D Shape Representation}
\label{sec:3.1}

Our \modelname represents shape using hybrid signed distance fields (SDF) as done by Wu \etal \cite{wu2023magicpony}, directly leveraging the benefits of explicit (\eg triangular meshes \cite{li2024fauna}) and implicit (\eg SDFs and radiance fields \cite{Antic_2025_ICCV, Deng_2021_CVPR, Pavllo_2023_CVPR}) 3D shape representation methods. More specifically, the object shape is represented implicitly using a neural field which is converted in real time into an explicit mesh by the {\it Differentiable Marching Tetrahedra} (DMTet \cite{e74-d_1_214}) method. As a result, our 3D shape representation is compact, supporting powerful deformation models, while allowing for topological changes. By leveraging the dense correspondences induced by a shared canonical template and its hybrid-SDF/DMTet-based deformation, \modelname can effectively learn and enforce semantic consistency during training. The canonical template mesh consisting of a canonical set of vertices is defined as follows:
\begin{equation} \label{eq:1}
    \mathcal{V}_{prior} = \{\mathbf{v}_i\}_{i=1}^{N}, \quad \mathbf{v}_i \in \mathbb{R}^3
\end{equation}
This template encodes category-level geometry and provides a shared reference frame across instances. The canonical template mesh is further deformed into a specific object instance $\mathcal{V}_{ins}$ to match the shape in each input image. 

\subsection{Semantic Consistency}
\label{sec:3.2}

The deformation network in our \modelname is represented as an MLP $\mathcal{F}_{def}: \mathbb{R}^3 \times \mathbb{R}^{d_f} \times \mathbb{R}^8 \to \mathbb{R}^3$, which predicts per-vertex offsets conditioned on canonical vertex coordinates $\mathcal{V}_{prior}$, global image features $\mathbf{f}_I$, and learned vertex index embeddings $\mathbf{e}_i$. Formally, let the canonical set of vertices be defined as in Equation~\ref{eq:1}. The deformation field predicts:
\begin{equation} \label{eq:4}
    \Delta \mathbf{v}_i = \mathcal{F}_{def} (\mathbf{v}_i, \mathbf{f}_I, \mathbf{e}_i)
\end{equation}
where the vertex index embeddings are defined as:
\begin{equation}
    \mathbf{e}_i = \mathcal{E}_\psi(i), \quad i \in \{1, ..., N\}, \quad \mathbf{e}_i \in \mathbb{R}^8
\end{equation}
The deformed vertex positions are then given by:
\begin{equation}
    \mathbf{v}'_i = \mathbf{v}_i + \Delta \mathbf{v}_i
\end{equation}
Finally, the deformed instance is represented as:
\begin{equation} \label{eq:7}
    \mathcal{V}_{ins} = \{\mathbf{v}'_i\}_{i=1}^{N}, \quad \mathbf{v}'_i \in \mathbb{R}^3
\end{equation}

By conditioning our deformation field on vertex index embeddings, \modelname encourages each canonical vertex to learn a persistent, identity-specific latent code, enabling consistent semantic correspondence across instances. More specifically, the embedding $\mathbf{e}_i = \mathcal{E}_\psi(i)$ assigns a fixed latent descriptor to the $i$-th canonical vertex, which is shared across all training examples. As a result, the deformation network $\mathcal{F}_{def}$ learns to associate each vertex index with a consistent semantic role (\eg chair leg tip, seat corner, backrest edge, \etc), rather than relying solely on local geometry. This discourages permutation ambiguity and arbitrary drift of vertices during optimization, and enforces cross-instance alignment in the canonical space. Consequently, vertices that occupy analogous semantic parts across different object instances are encouraged to undergo similar deformation patterns, promoting stable correspondence and semantically coherent shape reconstruction.

To further enforce semantic consistency, our \modelname also introduces a feature-level consistency loss aligning semantic features between the canonical mesh $\mathcal{V}_{prior}$ and deformed mesh $\mathcal{V}_{ins}$. For this purpose, we use a coordinate-based feature MLP, $\mathcal{F}_{feat}$, which maps each 3D vertex to a high-dimensional semantic embedding:
\begin{equation}
    f_i = \mathcal{F}_{feat}(\mathbf{v}_i), \quad \mathbf{v}_i \in \mathcal{V}_{prior} \cup \mathcal{V}_{ins}
\end{equation}
The feature-level consistency loss encourages corresponding vertices across canonical and deformed meshes to have similar embeddings:
\begin{equation}
    \mathcal{L}_{const} = \frac{1}{N}\sum^N_{i = 1} \left(f_i^{prior} - f_i^{ins}\right)^2
\end{equation}
This ensures that semantically corresponding vertices across different instances are aligned in feature space. More specifically, the semantic features of corresponding vertices are shared across instances.

\subsection{Geometric Aware Semantic Correspondence Retrieval}
\label{sec:3.3}

Our \modelname produces dense 3D correspondences across instances by representing each object category with a canonical mesh (Equation~\ref{eq:1}) and a learned deformation field (Equation~\ref{eq:4}). Each vertex $\mathbf{v}_i$ in the canonical mesh is semantically aligned with its deformed counterpart $\mathbf{v}'_i$ across all instances, providing a dense 3D vertex correspondence map:
\begin{equation}
\mathcal{C}_{3D}: \mathbf{v}_i \leftrightarrow \mathbf{v}'_i, \quad \forall i \in {1, \dots, N}.
\end{equation}

To extract 2D correspondences in a given image, we project the deformed 3D vertices of a given instance onto the image plane using the predicted pose $\theta$ and filter out self-occluded vertices using the rendered depth map $D$ via a z-buffer test:
\begin{equation}
\mathbf{u}_i = \Pi_\theta(\mathbf{v}'_i), \quad \mathbf{u}_i \in \mathbb{R}^2, \quad \text{s.t. } \mathbf{v}'_i \text{ is visible: } z_i = \mathbf{v}'^z_i \le D(\mathbf{u}_i)
\end{equation}
where $\Pi_\theta(\cdot)$ denotes the projection from 3D to 2D, and $z_i$ is the depth of the vertex in camera coordinates. For each 2D query keypoint $\mathbf{p}_q$ in the image, we first find the nearest projected visible vertex:
\begin{equation}
\hat{i} = \underset{i}{\arg\min} | \mathbf{p}_q - \mathbf{u}_i |_2
\end{equation}
To further refine these matches, we perform a local feature search around the nearest vertex in the target image:
\begin{equation} \label{eq:11}
\hat{\mathbf{p}}_q = \underset{\mathbf{p} \in \mathcal{N}(\mathbf{u}_{\hat{i}})}{\arg\max} \left\langle \mathbf{f}_I(\mathbf{p}_q), \mathbf{f}_I(\mathbf{p}) \right\rangle
\end{equation}
where $\mathcal{N}(\mathbf{u}_{\hat{i}})$ defines a local neighborhood around $\mathbf{u}_{\hat{i}}$, and $\langle \cdot, \cdot \rangle$ denotes the feature similarity. The resulting point $\hat{\mathbf{p}}_q$ gives a refined 2D correspondence aligned with the canonical 3D vertex.

\begin{figure}
    \centering
    \includegraphics[width=\linewidth]{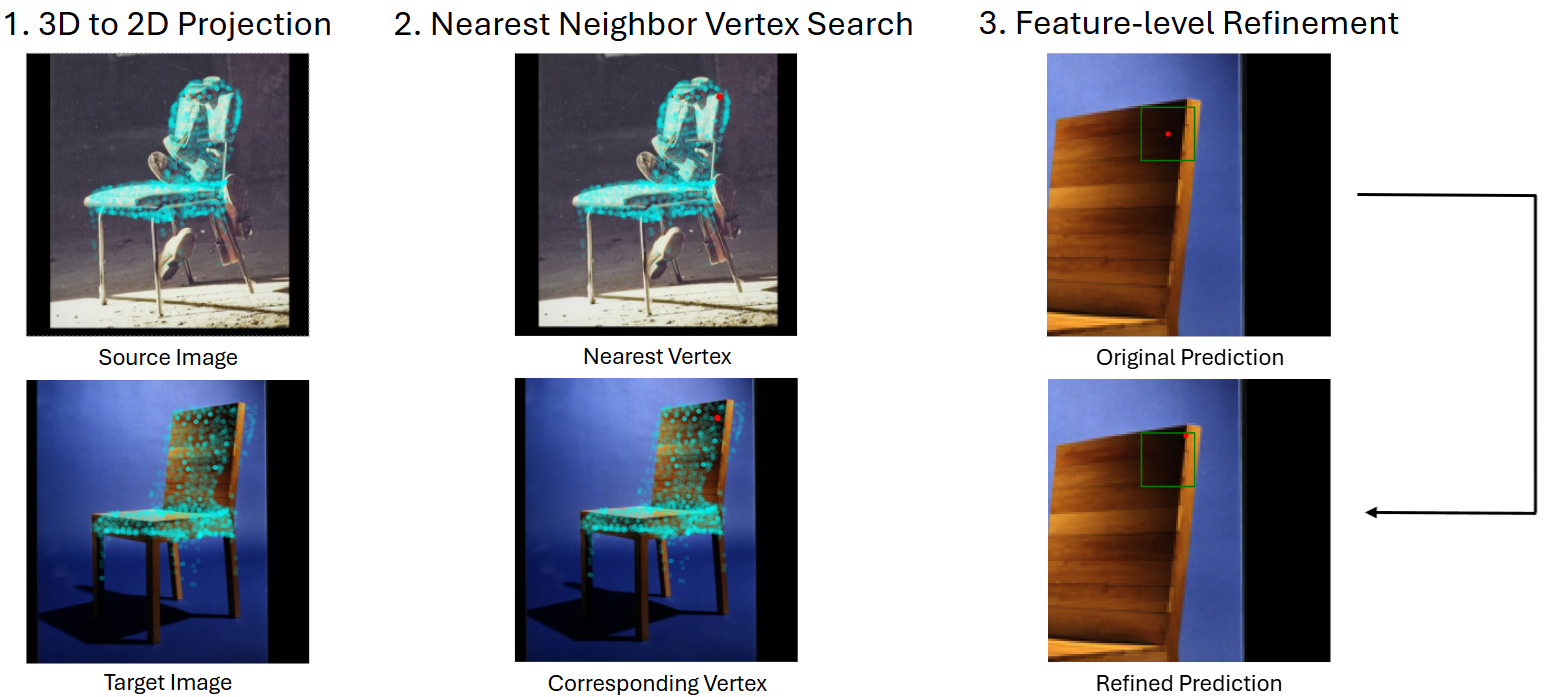}
    \caption{\textbf{Geometric Aware Semantic Correspondence Retrieval} To obtain 2D correspondences from our learned dense 3D representation, we first project canonical vertices into the image plane. Given a source keypoint, we identify the nearest visible vertex in 2D as its corresponding 3D anchor. This vertex is then mapped to the target image via the shared canonical space, producing an initial 2D prediction. Finally, we refine the prediction by performing a local feature search in the target image to select the most semantically consistent match.}
    \label{fig:evaluation}
\end{figure}

By explicitly coupling 3D geometry with 2D image features, our \modelname is able to establish semantically meaningful correspondences across instances. The learned deformation field provides dense 3D vertex correspondences between the canonical mesh and each object instance, while the predicted pose and depth map allow us to project these vertices onto the image plane. In turn, local image features extracted at these projected locations are used to refine the 2D correspondences, enabling the model to reconcile geometric alignment with appearance cues via the local feature search. This combination ensures that the resulting correspondences are both geometrically accurate and semantically consistent, even across instances with large intra-category variation or partial occlusions (Figure~\ref{fig:evaluation}).

\subsection{Vision Foundation Models as Supervision}
\label{sec:3.4}
Recent deformable 3D reconstruction methods achieve strong results on articulated categories by leveraging consistent topology and constrained articulation priors \cite{SMPL:2015, Qian_2024_CVPR, wu2023magicpony, li2024fauna}. In contrast, category-level reconstruction of rigid objects presents different challenges due to the lack of explicit articulation and large intra-category geometric variation, which weakens shared deformation priors and complicates semantic alignment across instances.

To mitigate these challenges, we incorporate additional geometric supervision from off-the-shelf estimators during training. Depth supervision constrains surface geometry and discourages degenerate deformations, while explicit pose supervision resolves ambiguities caused by strong object symmetries and large viewpoint variation by decoupling pose from shape deformation. Specifically, we supervise rendered depth using a shift-and-scale invariant L2 loss \cite{ranftl2020towards}:
\begin{equation}
\mathcal{L}_{depth} = \mathds{1}_{mask}\left|(\alpha D_{pred} + \beta) - D_{gt}\right|_2 ,
\end{equation}
where $\alpha$ and $\beta$ are estimated via least squares.

We additionally supervise pose using a cosine regression loss:
\begin{equation}
\mathcal{L}_{pose} = 1 - \cos(\theta_{pred} - \theta_{gt}) ,
\end{equation}
which stabilizes training by preventing pose ambiguity from being absorbed into the deformation field.

While these geometric losses improve reconstruction fidelity and training stability, they do not by themselves enforce semantic alignment across instances, motivating the semantic consistency mechanisms introduced in our model.

\section{Experiments}
\label{sec:experiments}

We conduct extensive experiments and evaluate the effectiveness of \modelname across multiple object categories and downstream tasks. Our evaluation focuses on cross-instance semantic correspondence, highlighting the benefits of our geometry-aware canonical representation. We describe our experimental setup in Section~\ref{sec:4.1}. Then, in Section~\ref{sec:4.2}, we demonstrate that our model improves over prior 3D deformable models in semantic correspondence. Next, in Section~\ref{sec:4.3}, we present quantitative 3D reconstruction quality results. Additionally, we provide a comprehensive qualitative comparison and discussion between our model and baselines in Section~\ref{sec:4.4}. Lastly, we provide ablations to analyze each component of our \modelname in Section~\ref{sec:4.5}.

\subsection{Experimental Details}
\label{sec:4.1}

\textbf{Training Dataset.} We train \modelname using the in-the-wild images of the ImageNet split from the PASCAL3D+ \cite{xiang2014beyond} dataset. Following prior deformable reconstruction works \cite{wu2023magicpony, Pavllo_2023_CVPR}, we train a separate category-specific model for each object class. The dataset provides diverse images for 12 object categories with variation in pose, scale, and appearance, enabling the model to learn a robust canonical representation and deformation field. Each image is supplemented by object masks using Segment Anything Model \cite{kirillov2023segment}, depth maps using Depth Anything V2 \cite{yang2024depth}, and estimated object poses using Orient Anything V2 \cite{wangorient}.

\textbf{Evaluation on In-the-Wild Datasets.} To evaluate \modelname on in-the-wild data, we use SPair-71k \cite{min2020spair71k} which provides keypoint annotations for semantic correspondence using image pairs derived from the PASCAL split of PASCAL3D+. For comprehension, we conduct semantic correspondence evaluation on all 10 rigid object categories shared between SPair-71k and PASCAL3D+. Notably, there exists a substantial distribution gap between the PASCAL3D+ ImageNet split used for training and the PASCAL split used for evaluation. More specifically, the PASCAL split contains more challenging images with uncommon viewpoints, stronger occlusions, and truncated objects. To reduce the impact of this gap and ensure fair evaluation, we restrict our experiments to non-truncated object instances for both instance segmentation and semantic correspondence. All baseline methods are evaluated on the same subset for a consistent and fair comparison.

\textbf{Implementation Details.} Our global image encoder is a DINOv2-ViT-S model \cite{dosovitskiy2020image} encoder, providing semantic-rich embeddings for alignment with the canonical mesh. The deformation field $\mathcal{F}_{def}$ is modeled by a 5-layer coordinate-based MLP that predicts per-vertex 3D offsets from the canonical template using vertex coordinates, image features, and learnable vertex embeddings. The pose network takes DINOv2 patch features as input and predicts instance-specific 3D object orientation using a small convolutional encoder that reduces 32×32 patch embeddings to a latent pose vector. We utilize the rasterizer proposed by \cite{laine2020modular} for efficient differentiable rendering. During training, we optimize for losses following \cite{wu2023magicpony, li2024fauna} along with the losses introduced in Sections \ref{sec:3.2} and \ref{sec:3.4}.

\subsection{Semantic Correspondence}
\label{sec:4.2}

We evaluate the effectiveness of \modelname on semantic correspondence in the top half of Table~\ref{tab:pck}. We compare against MagicPony (MP)~\cite{wu2023magicpony}, a DINOv2 feature matching baseline, and a DINOv2~\cite{oquab2023dinov2} PCA baseline. We choose MagicPony as a baseline because it is a strong category-level 3D deformable model trained without semantic correspondence as their primary objective. To adapt MagicPony for rigid objects, we remove the articulation head during training. At evaluation time, we project the reconstructed 3D mesh into the image plane to obtain correspondences. To isolate the effect of semantic consistency, we additionally augment MagicPony with the same pose and depth supervision (Section~\ref{sec:3.4}) used in our model.

For the DINOv2 PCA model, DINOv2 features are extracted from the ImageNet split of PASCAL3D+ and reduced to 16 dimensions using PCA. We include this baseline because the same DINOv2 features are used in the local feature matching step described in Section~\ref{sec:3.3}, allowing us to isolate the contribution of our geometry-aware formulation. For correspondence estimation, the local search region in Equation~\ref{eq:11} is restricted to a $24 \times 24$ pixel window (less than 1\% of the $256 \times 256$ image resolution).

On average, \modelname improves PCK\texttt{@}0.1 by 6.7 points over the DINOv2 PCA baseline and 10.3 points over the DINOv2 baseline, demonstrating that coupling semantically consistent 3D geometry with 2D image features substantially enhances matching accuracy beyond feature similarity alone. Compared to MagicPony, \modelname improves PCK\texttt{@}0.1 from 37.4 to 52.1, a gain of 14.7 points. This result reveals a key limitation of reconstruction-focused deformable models: despite learning accurate geometry, these models fail to learn semantically consistent canonical representations.

For fair comparison, adding Depth Anything V2 supervision improves MagicPony from 37.4 to 38.6 PCK\texttt{@}0.1, while Orient Anything V2 further improves performance to 43.9. Even under identical visual foundation model supervision, our method outperforms MagicPony by 8.2 PCK\texttt{@}0.1, highlighting the contribution of our semantic consistency formulation.

\begin{table}[h]
  \caption{In the top section, \textbf{evaluation of PCK\texttt{@}0.1} on all 10 rigid object categories of SPair-71k. In the bottom section, \textbf{ablation} on our model design choices. Each variant removes a single component while keeping the rest of the model unchanged.}
  \label{tab:pck}
  \centering
  \setlength{\tabcolsep}{4pt}
  \begin{tabular}{lccccccccccc}
    \toprule
    & 
    \raisebox{-0.25\height}{\includegraphics[height=2.0em]{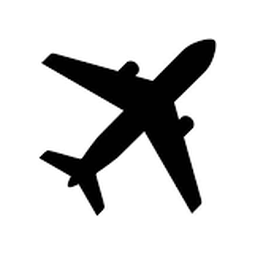}} &
    \raisebox{-0.25\height}{\includegraphics[height=2.0em]{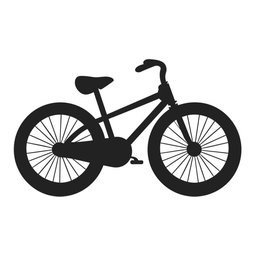}} &
    \raisebox{-0.25\height}{\includegraphics[height=2.0em]{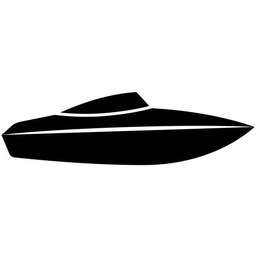}} &
    \raisebox{-0.25\height}{\includegraphics[height=2.0em]{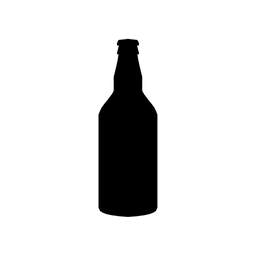}} &
    \raisebox{-0.25\height}{\includegraphics[height=2.0em]{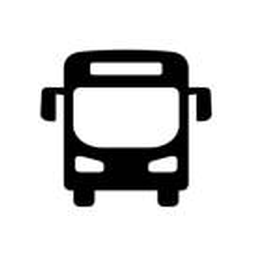}} &
    \raisebox{-0.25\height}{\includegraphics[height=2.0em]{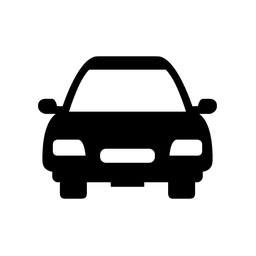}} &
    \raisebox{-0.25\height}{\includegraphics[height=2.0em]{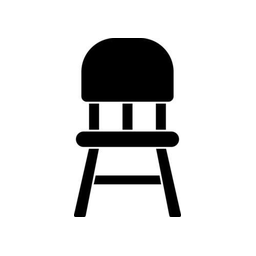}} &
    \raisebox{-0.25\height}{\includegraphics[height=2.0em]{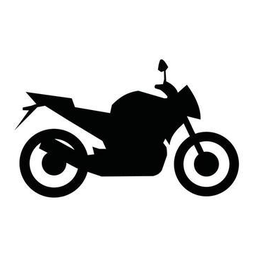}} &
    \raisebox{-0.25\height}{\includegraphics[height=2.0em]{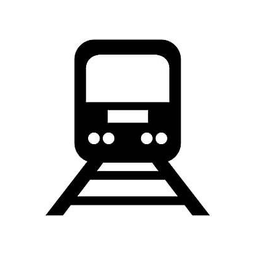}} &
    \raisebox{-0.25\height}{\includegraphics[height=2.0em]{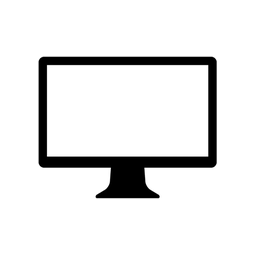}} &
    \raisebox{0.2em}{AVG} \\
    \midrule
    MP & 60.6 & 45.1 & 22.6 & 40.3 & 23.4 & 33.9 & 27.6 & 47.4 & 32.6 & 40.6 & 37.4 \\
    MP + Depth & 60.9 & 46.1 & 25.8 & 40.5 & 23.9 & 35.1 & 30.0 & 49.3 & 32.9 & 41.4 & 38.6 \\
    MP + Pose & 61.2 & 50.9 & 33.0 & 42.7 & 41.8 & 44.5 & 31.6 & 51.3 & 37.8 & 43.8 & 43.9   \\
    DINOv2 & 62.8 & 47.8 & 30.2 & 41.1 & 43.1 & 43.6 & 30.2 & 48.6 & 45.8 & 25.1 & 41.8 \\
    DINOv2 PCA    & \textbf{70.5} & \textbf{69.0} & 39.3 & 36.1 & 41.0 & 48.2 & 31.0 & \textbf{56.2} & \textbf{46.5} & 15.8 & 45.4 \\
    Ours      & 62.9 & 57.6 & \textbf{42.9} & \textbf{45.0} & \textbf{62.4} & \textbf{63.5} & \textbf{37.6} & 54.6 & 45.3 & \textbf{49.1} & \textbf{52.1} \\
    \midrule
    w/o ind. emb. & 61.8 & 56.5 & 42.3 & 44.6 & 62.0 & 61.7 & 37.0 & 54.3 & 44.8 & 48.6 & 51.4 \\
    w/o feat. cons. & 62.0 & 56.3 & 42.5 & 44.7 & 59.5 & 61.3 & 37.2 & 54.4 & 45.0 & 48.8 & 51.2 \\
    w/o \ref{sec:3.2} (both)      & 61.3 & 56.1 & 41.4 & 43.5 & 61.4 & 60.9 & 36.7 & 54.2 & 43.6 & 47.9 & 50.7 \\
    w/o depth      & 62.8 & 56.3 & 38.6 & 43.8 & 58.6 & 62.0 & 35.7 & 52.6 & 44.9 & 48.2 & 50.4 \\
    w/o pose      & 61.1 & 50.4 & 32.3 & 42.0 & 44.7 & 49.9 & 31.5 & 50.2 & 39.2 & 45.7 & 44.7 \\
    \bottomrule
  \end{tabular}
\end{table}

\subsection{3D Reconstruction Quality}
\label{sec:4.3}

While \modelname is designed to improve semantic consistency, we additionally evaluate 3D reconstruction quality in Table~\ref{tab:3d}. We align the predicted meshes with PASCAL3D+ CAD models using ICP~\cite{besl1992method} and compute the normalized Chamfer distance between sampled surface points. \modelname reduces normalized Chamfer distance from 8.2 (MagicPony) to 6.2, a 24\% relative improvement, indicating that enforcing semantic consistency also improves geometric fidelity.

\begin{table}[h]
  \caption{\textbf{Evaluation of normalized Chamfer distance} on all 10 rigid object categories of PASCAL3D+.}
  \label{tab:3d}
  \centering
  \setlength{\tabcolsep}{4pt}
  \begin{tabular}{lccccccccccc}
    \toprule
    & 
    \raisebox{-0.25\height}{\includegraphics[height=2.0em]{aeroplane_cropped.png}} &
    \raisebox{-0.25\height}{\includegraphics[height=2.0em]{bicycle_cropped.png}} &
    \raisebox{-0.25\height}{\includegraphics[height=2.0em]{boat_cropped.png}} &
    \raisebox{-0.25\height}{\includegraphics[height=2.0em]{bottle_cropped.png}} &
    \raisebox{-0.25\height}{\includegraphics[height=2.0em]{bus_cropped.png}} &
    \raisebox{-0.25\height}{\includegraphics[height=2.0em]{car_cropped.png}} &
    \raisebox{-0.25\height}{\includegraphics[height=2.0em]{chair_cropped.png}} &
    \raisebox{-0.25\height}{\includegraphics[height=2.0em]{motorbike_cropped.png}} &
    \raisebox{-0.25\height}{\includegraphics[height=2.0em]{train_cropped.png}} &
    \raisebox{-0.25\height}{\includegraphics[height=2.0em]{tvmonitor_cropped.png}} &
    \raisebox{0.2em}{AVG} \\
    \midrule
    MagicPony & 7.3 & 8.7 & 5.0 & \textbf{2.5} & 16.3 & 5.6 & 5.2 & 6.6 & 19.2 & 5.5 & 8.2 \\
    Ours & \textbf{6.5} & \textbf{8.1} & \textbf{2.6} & 2.7 & \textbf{9.6} & \textbf{5.1} & \textbf{4.4} & \textbf{5.8} & \textbf{14.9} & \textbf{2.2} & \textbf{6.2} \\
    \bottomrule
  \end{tabular}
\end{table}

\subsection{Qualitative Comparisons}
\label{sec:4.4}

\begin{figure}
    \centering
    \includegraphics[width=0.60\linewidth]{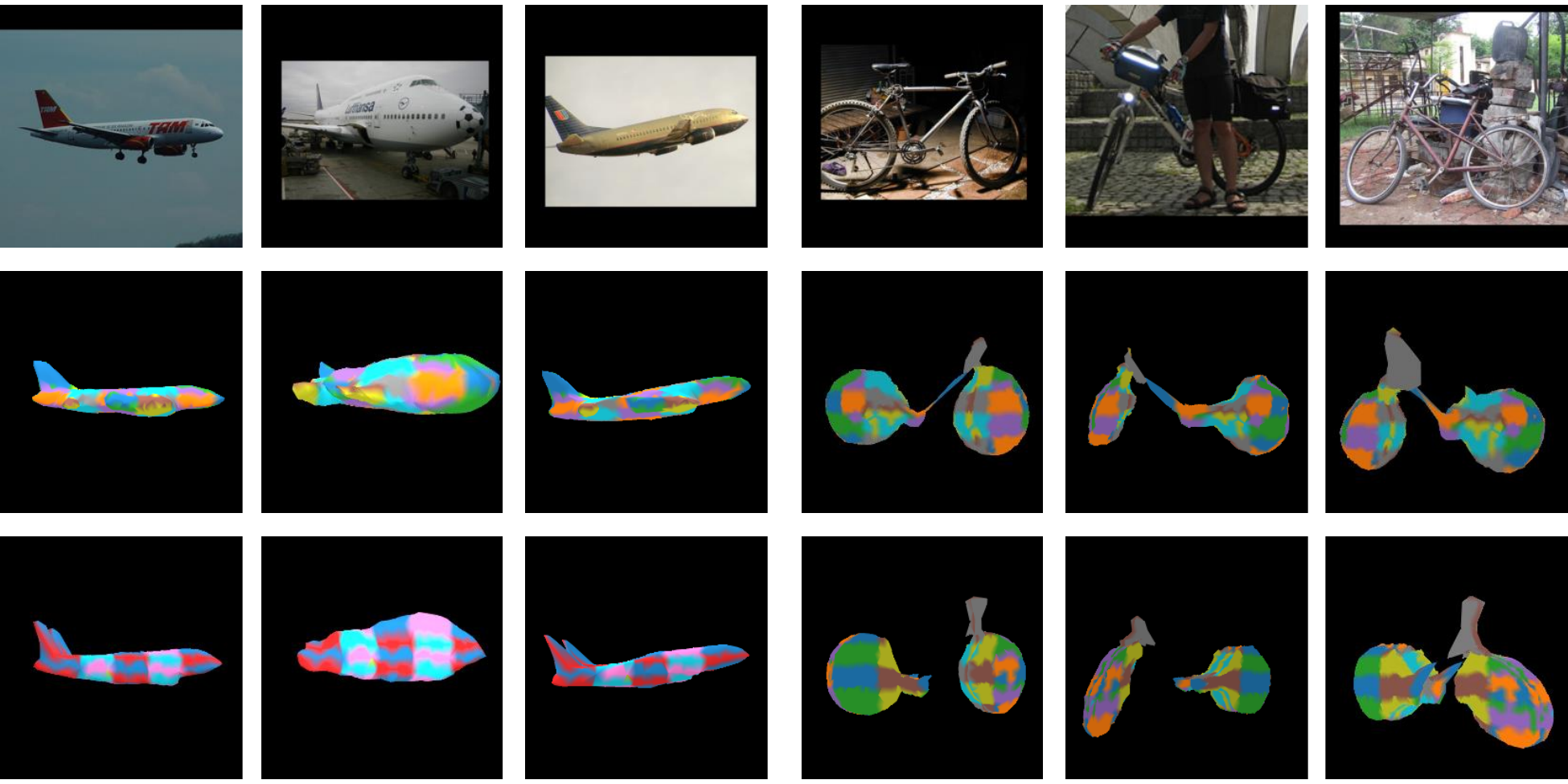}
    \includegraphics[width=0.60\linewidth]{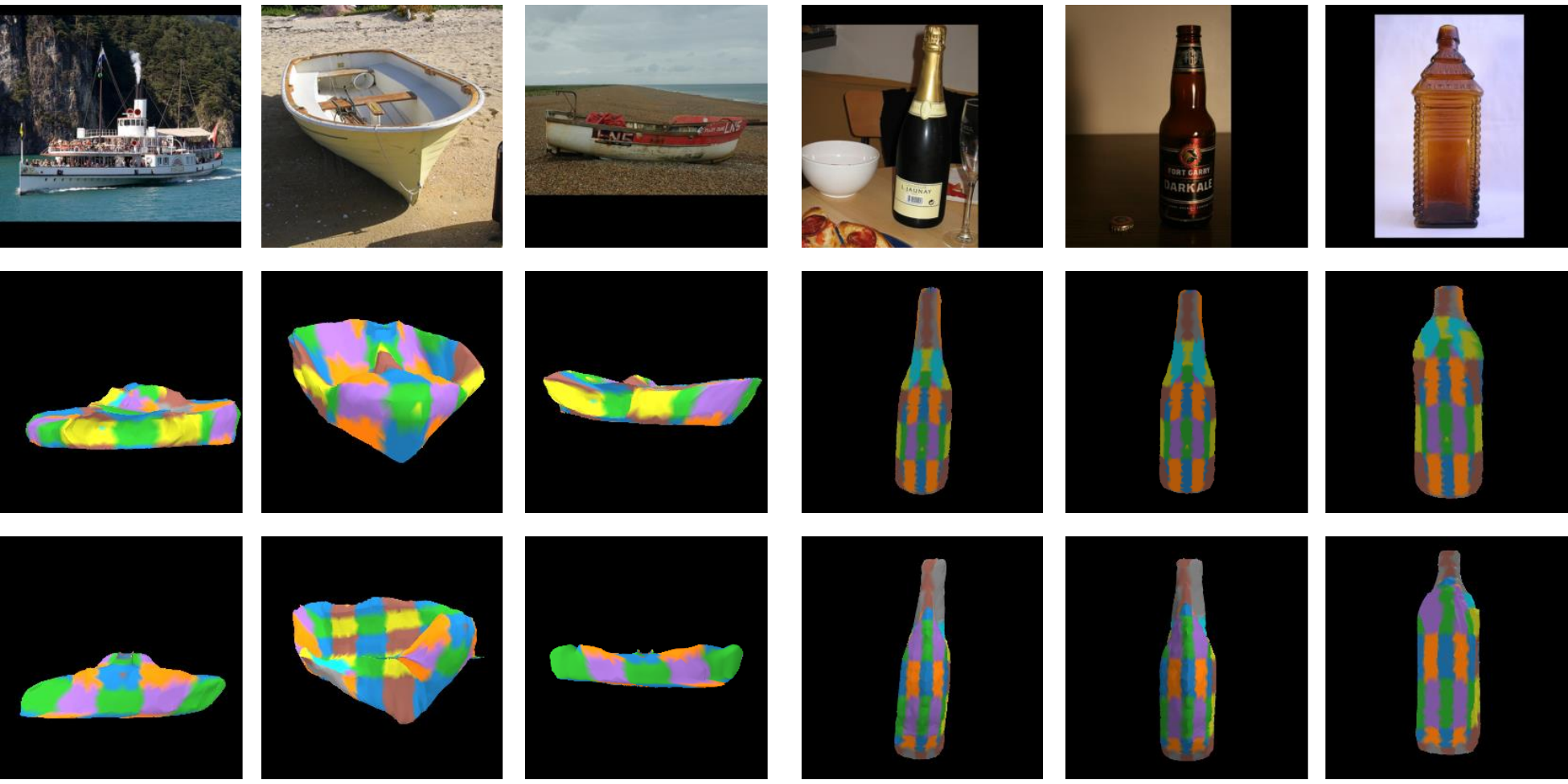}
    \caption{\textbf{Qualitative Comparison with Baseline.} The second row visualizes dense correspondences produced by our method, while the third row shows results from MagicPony~\cite{wu2023magicpony}. Our approach yields more semantically consistent correspondences, reflected by coherent color patterns on analogous parts, whereas MagicPony exhibits inconsistent color assignments across instances. For example, our method shows a consistent brown and yellow for the aeroplane wings, whereas the wings of MagicPony are dark blue in the first image, but light blue in the third. For boats, our model learns a consistent hull as shown by the color patterns, however, MagicPony is less consistent with a gray patch in the second image.}
    \label{fig:qualitative_a}
\end{figure}

\begin{figure}
    \centering
    \includegraphics[width=0.80\linewidth]{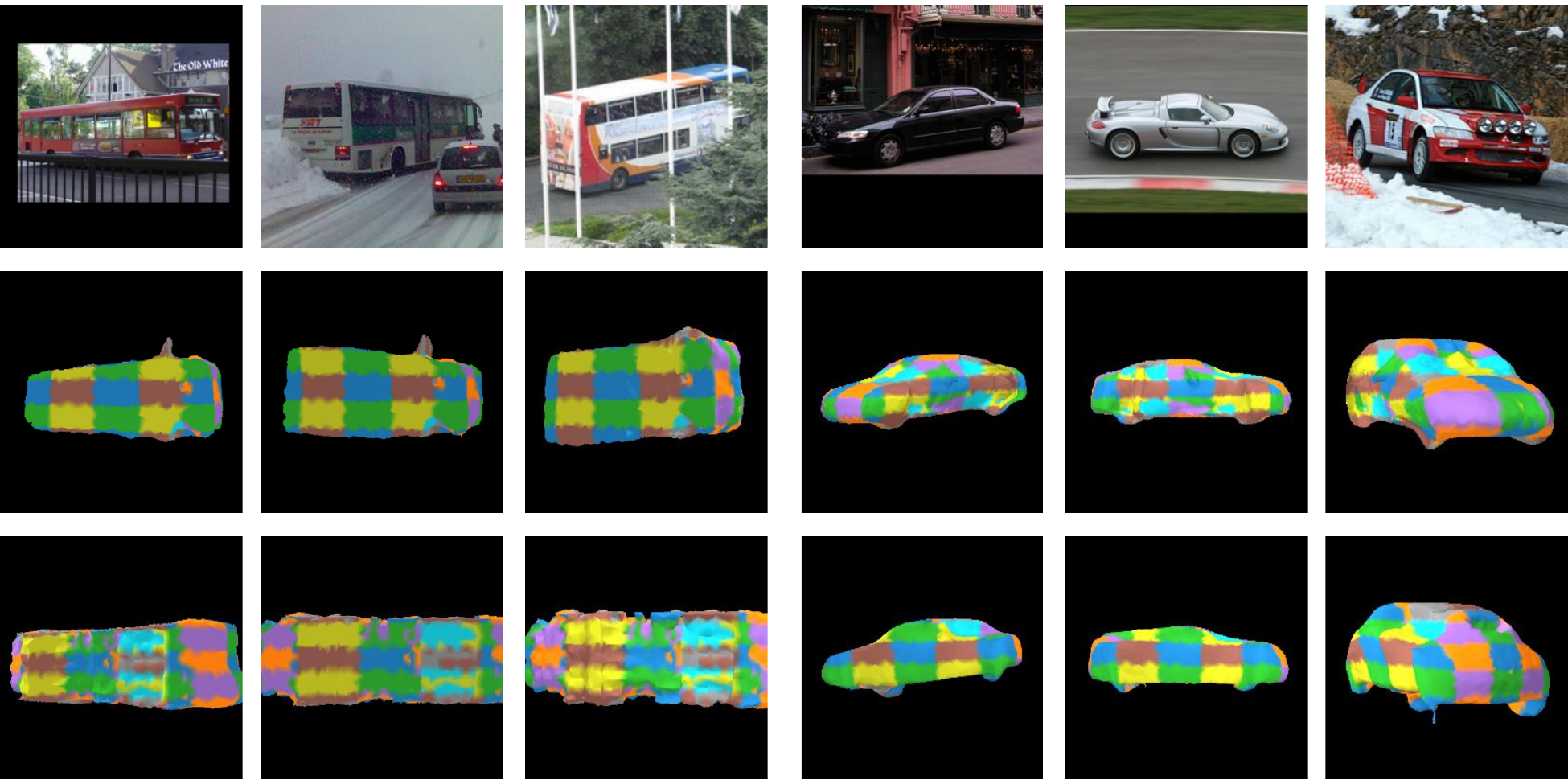}
    \includegraphics[width=0.80\linewidth]{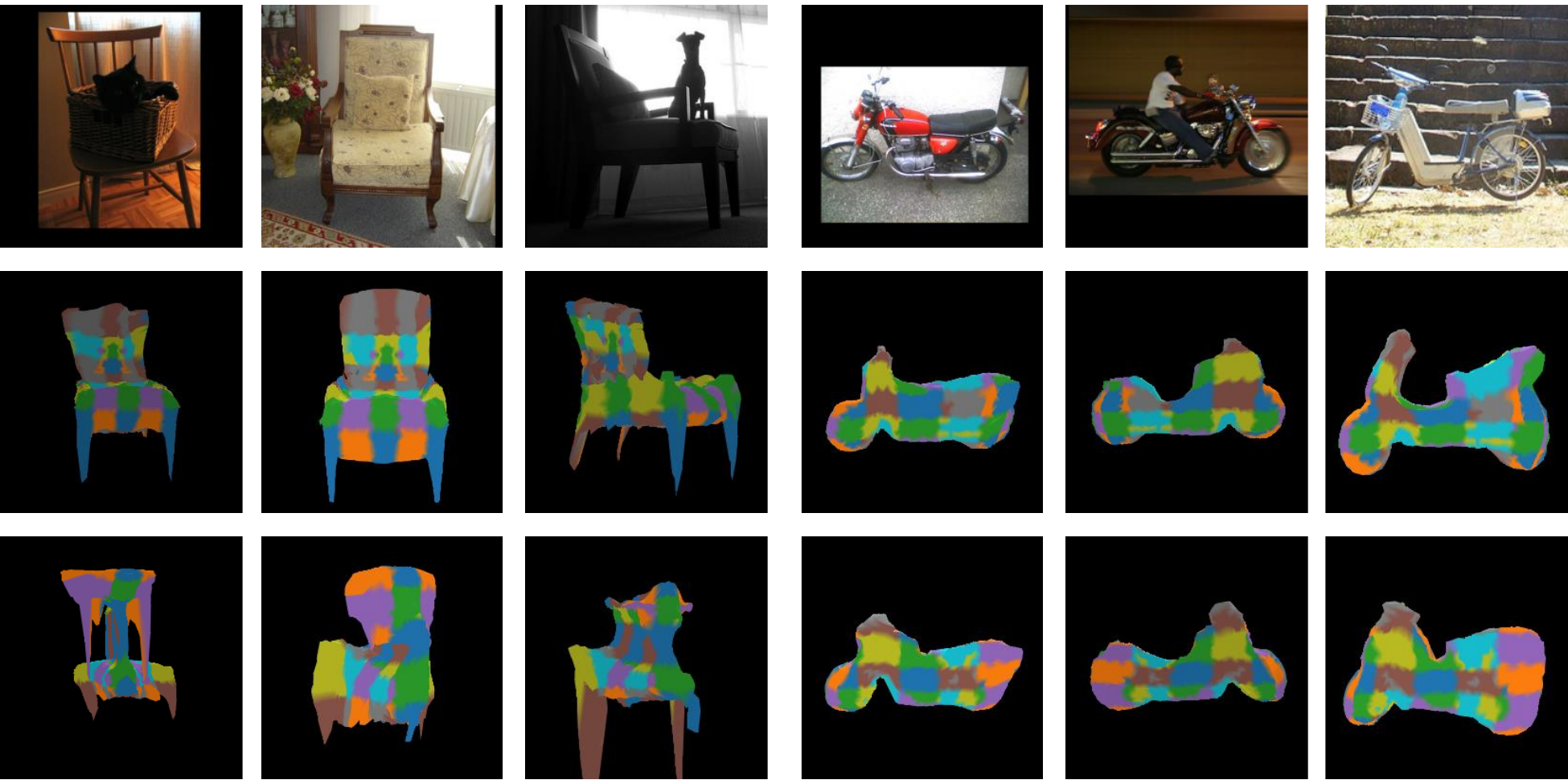}
    \includegraphics[width=0.80\linewidth]{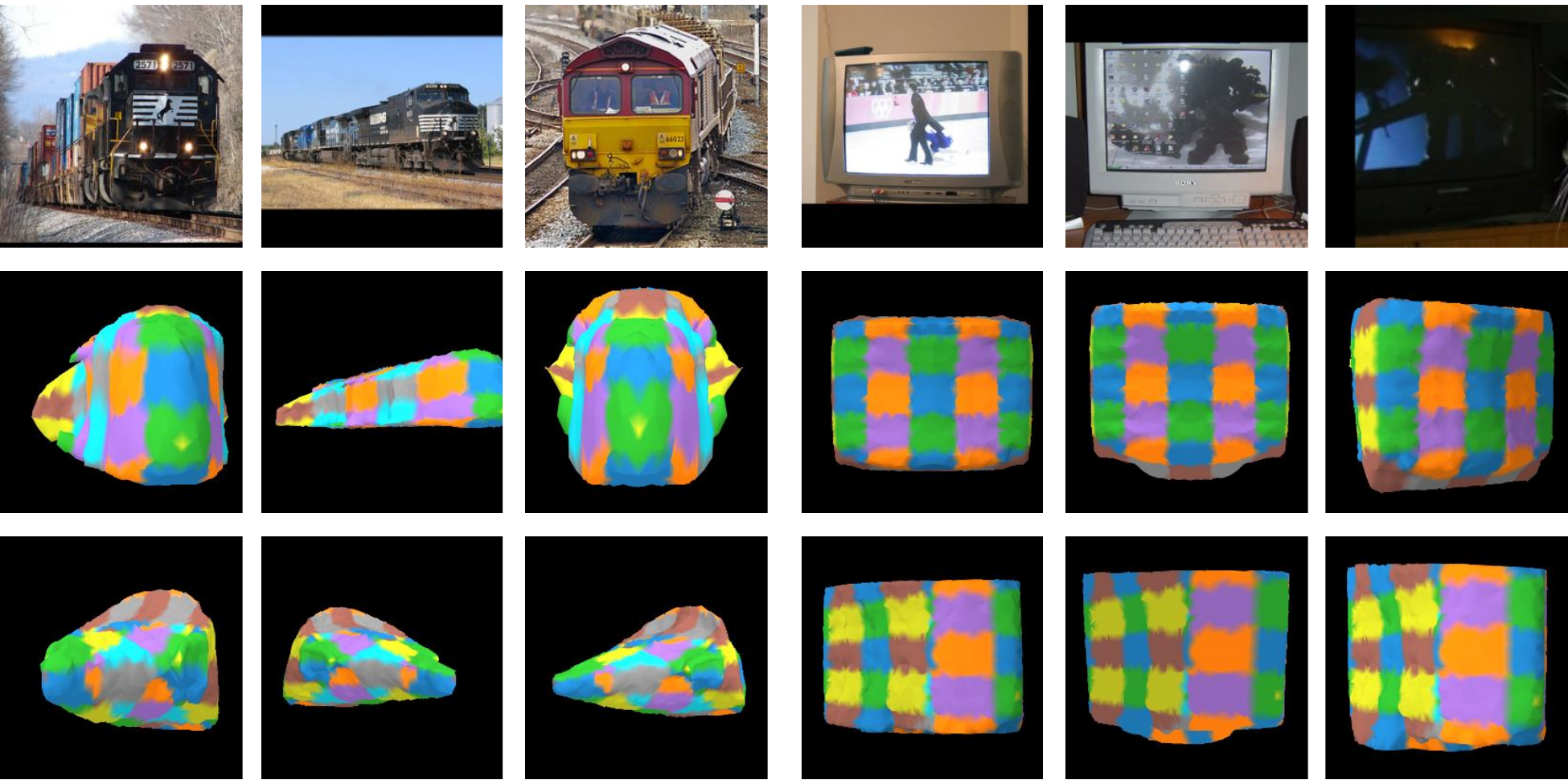}
    \caption{\textbf{Additional Qualitative Comparison with Baseline.} The second row visualizes dense correspondences produced by our method, while the third row shows results from MagicPony~\cite{wu2023magicpony}. Our approach yields more semantically consistent correspondences, reflected by coherent color patterns on analogous parts, whereas MagicPony exhibits inconsistent color assignments across instances. For example, for cars, the first image of MagicPony has a green patch for the windshield, but the other two images have a cyan color. For chairs, while MagicPony reconstruction is not great, the chair legs are not consistent with the second chair having a blue leg, while the others having brown legs.}
    \label{fig:qualitative_b}
\end{figure}

We present structured qualitative comparisons between \modelname and MagicPony. For all examples, methods are evaluated under identical conditions, and colors indicate vertex identities in the learned canonical space, transferred across instances. Figure~\ref{fig:qualitative_a} shows aeroplane, bicycle, boat, and bottle categories, while Figure~\ref{fig:qualitative_b} presents bus, car, chair, motorbike, train, and tv monitor. Across all categories, \modelname produces correspondences that are semantically stable across instances, as evidenced by coherent color patterns assigned to analogous object parts (\eg wings, hulls, seats, and wheels). In contrast, MagicPony frequently exhibits semantic drift: regions corresponding to the same semantic part are assigned different canonical vertices across instances, resulting in inconsistent color assignments under changes in shape, viewpoint, or occlusion.

These qualitative results complement the quantitative gains in Table~\ref{tab:pck}, illustrating that our improvements arise from increased semantic consistency in the learned canonical space rather than from reconstruction fidelity alone.

\paragraph{Category-wise Analysis.}
We observe consistent improvements across categories, with particularly strong gains for objects exhibiting symmetry, or thin structures.

\textbf{Aeroplane.} Aeroplanes contain symmetric and repetitive parts such as wings and tail surfaces, which often lead to ambiguous canonical alignment. MagicPony frequently assigns different canonical vertices to the same wing regions across instances. In contrast, our method consistently aligns corresponding wing tips, resulting in stable cross-instance correspondence.

\textbf{Boat.} Boats exhibit substantial variation in hull shape and superstructure, making purely geometry-driven alignment unstable. MagicPony often produces fragmented or drifting correspondences along the hull. Our method learns a coherent hull representation, preserving semantic consistency even under viewpoint changes.

\textbf{Bicycle.} Thin structures such as spokes, frames, and handlebars are prone to correspondence noise. Our method produces more stable alignments for these components by anchoring deformation to a semantically consistent canonical space.

\textbf{Bus and TV Monitor.} These categories exhibit strong rotational symmetry, leading to pose-related ambiguities when trained without pose supervision. While MagicPony shows inconsistent correspondence under rotation, our explicit pose supervision combined with semantic constraints results in stable alignment of planar regions.

\textbf{Chair.} Chairs vary significantly in topology (\eg number of legs, presence of armrests), which challenges shared deformation priors. MagicPony mismatches legs across instances. By preserving vertex identity, our method maintains consistent correspondences for functional parts such as leg endpoints.

Overall, these examples demonstrate that enforcing semantic consistency during deformation learning leads to robust correspondence across diverse object categories, even when reconstruction quality alone is insufficient to constrain semantic alignment.

\subsection{Ablation Studies}
\label{sec:4.5}

In the bottom half of Table~\ref{tab:pck}, we present ablations of our design choices. Each variant removes a component while keeping the rest of the model unchanged. We observe that conditioning the deformation network on vertex indices and incorporating the feature-level consistency loss improve semantic consistency. These improvements are consistent across all ten categories. Similarly, removing depth supervision also slightly degrades model performance, demonstrating that both geometric and semantic regularization both play an important role in learning stable correspondence. The most significant drop occurs when removing pose supervision, highlighting that strong pose is a necessary prerequisite for category-level correspondence under large viewpoint variation. However, pose alone is insufficient to explain our gains: even with pose supervision retained, removing semantic consistency components leads to clear and consistent performance drops. This indicates that our proposed semantic consistency mechanisms are still responsible for improving the quality and robustness of dense correspondences across object instances.

\section{Conclusion}
We have presented \modelname, a framework for learning semantically consistent deformable 3D representations from in-the-wild images. Unlike traditional deformable reconstruction methods that prioritize visual fidelity, \modelname explicitly enforces semantic consistency across object instances through vertex-index–conditioned deformation and feature-level alignment. By combining a canonical mesh with instance-specific deformations guided by semantic image features, our approach produces dense, stable correspondences across diverse object categories, even under large intra-class variation and partial occlusion.

Experimental results demonstrate that \modelname significantly improves semantic correspondence over existing deformable models and 2D feature-based baselines, establishing that deformable 3D representations can serve as robust semantic priors rather than merely reconstruction tools. Beyond semantic correspondence, our method provides a flexible foundation for downstream tasks requiring part-level 3D understanding, including object manipulation, robotics, augmented reality, and category-level reasoning.

\begin{comment}
    \section*{Acknowledgments}
    Please insert your acknowledgments here.
\end{comment}

\clearpage

\bibliographystyle{splncs04}
\bibliography{main}

\clearpage
\setcounter{section}{0}
\renewcommand{\thesection}{\Alph{section}}

\section{Limitations}

Similar to other deformable 3D reconstruction methods such as \cite{wu2023magicpony}, \modelname relies on relatively clean training data. In particular, the input images should contain largely unoccluded, non-truncated objects, as heavy occlusion or truncation can degrade the quality of the learned geometry and correspondences. Additionally, extreme out-of-distribution viewpoints causing incorrect Orient Anything V2 predictions may propagate errors into the deformation field during training. Lastly, we also observe a distribution gap between the training and evaluation data, which can result in lower pose estimation accuracy when test images exhibit viewpoints or appearance variations not well represented during training. This can potentially be addressed by training on a larger and more diverse dataset.

\section{Qualitative Examples on SPair-71k}

\begin{figure}[t]
    \centering
    \includegraphics[width=0.49\linewidth]{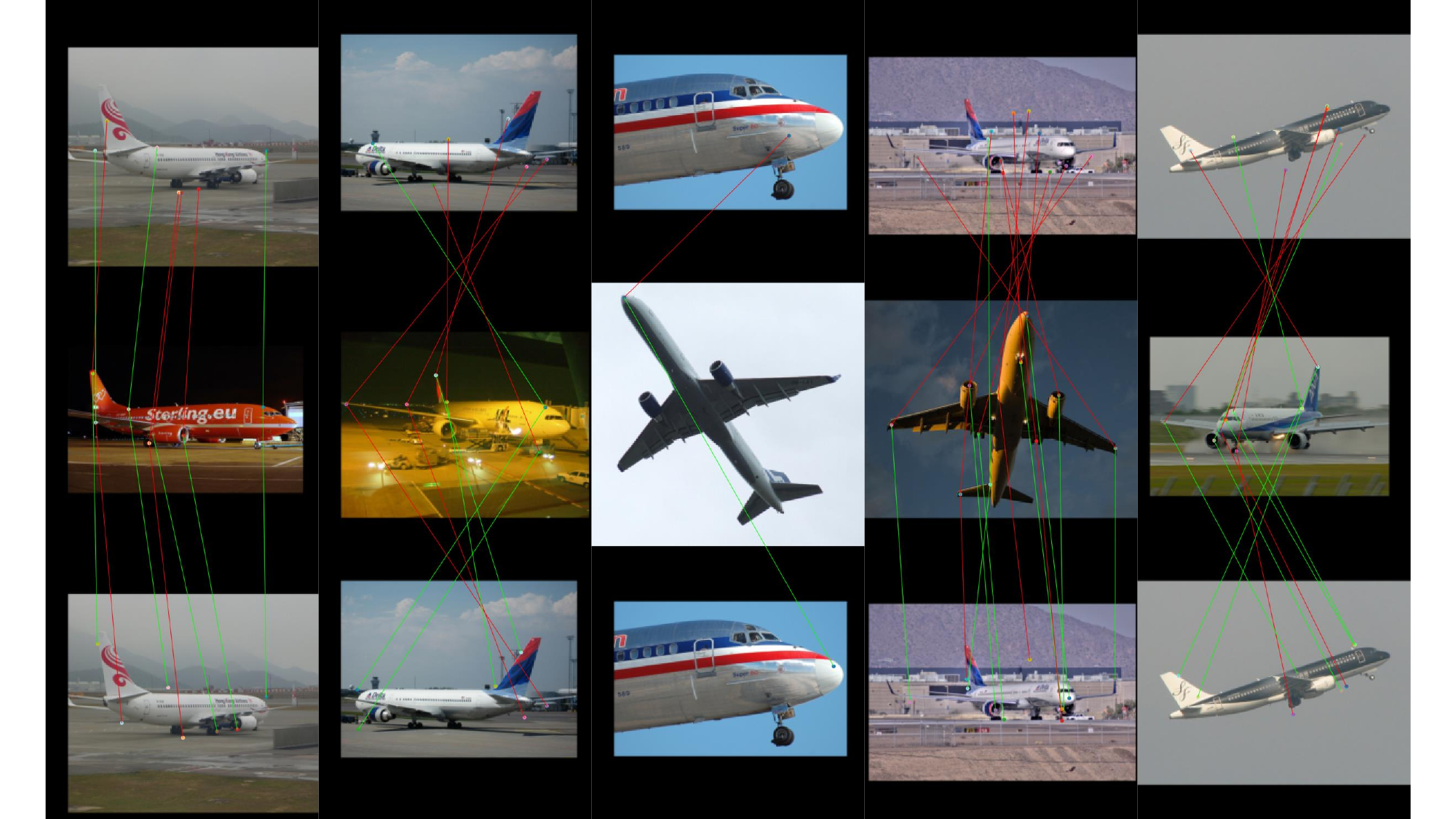}
    \hspace{0.001\linewidth}
    \includegraphics[width=0.49\linewidth]{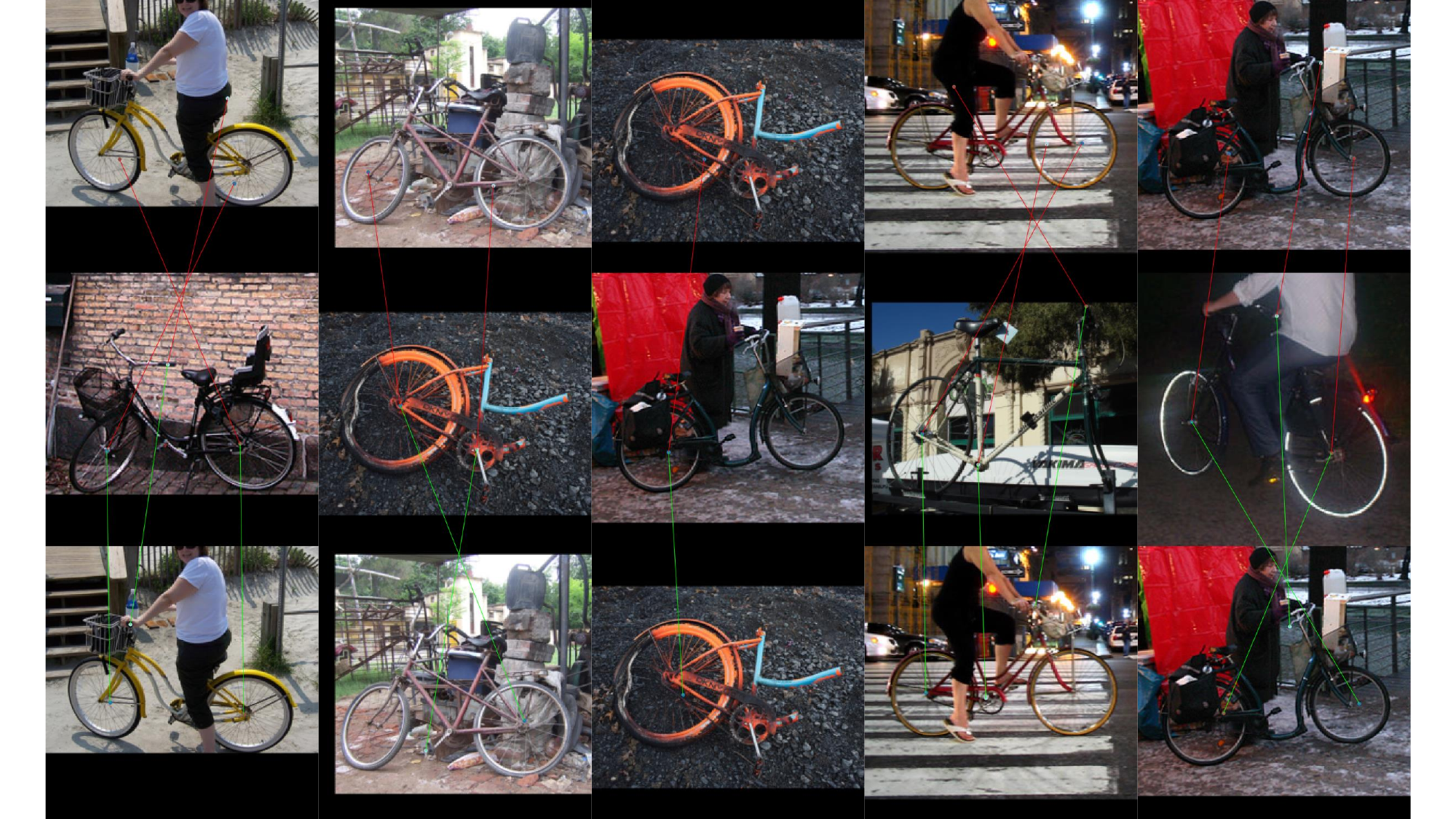}
    \includegraphics[width=0.49\linewidth]{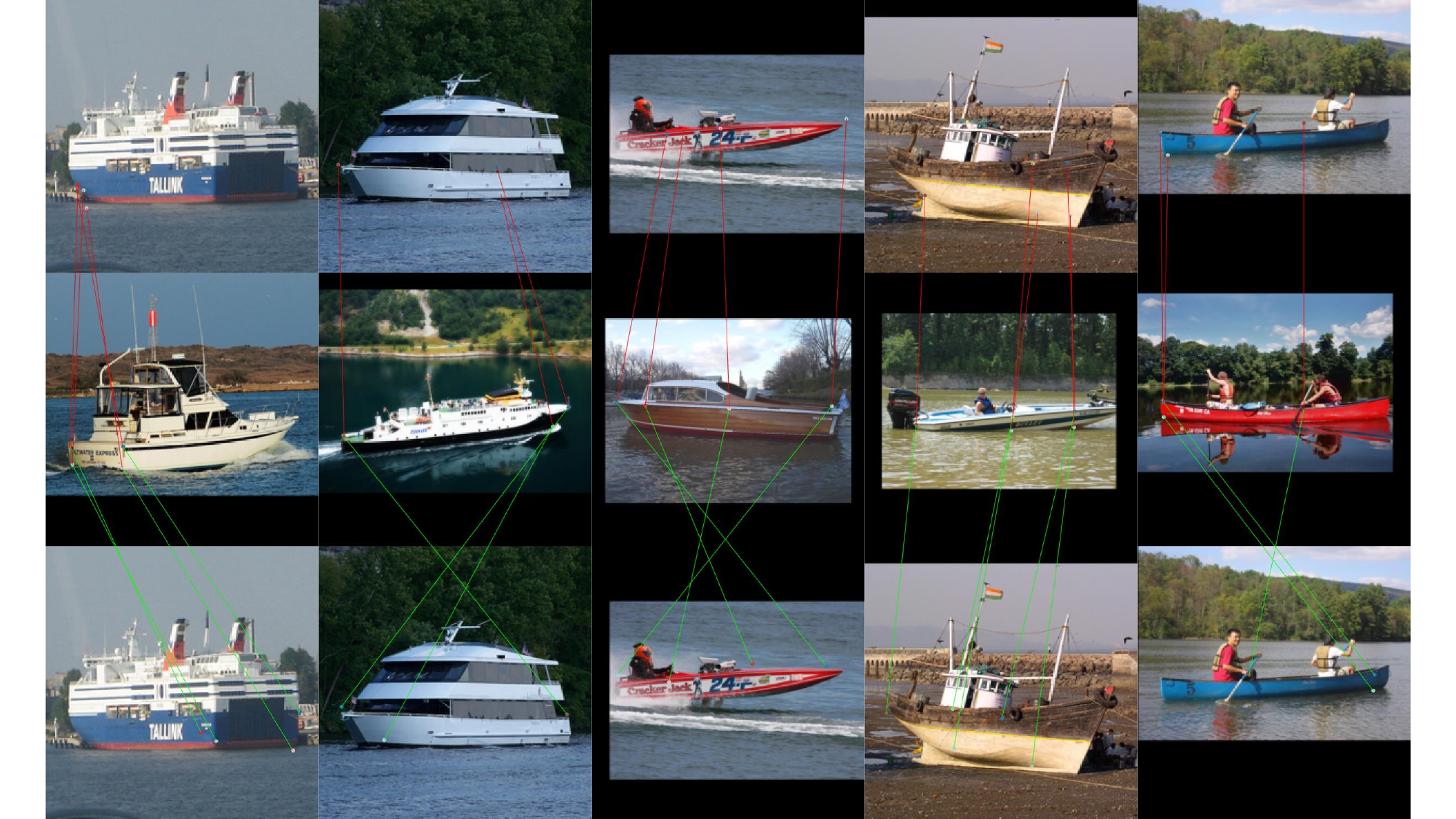}
    \hspace{0.001\linewidth}
    \includegraphics[width=0.49\linewidth]{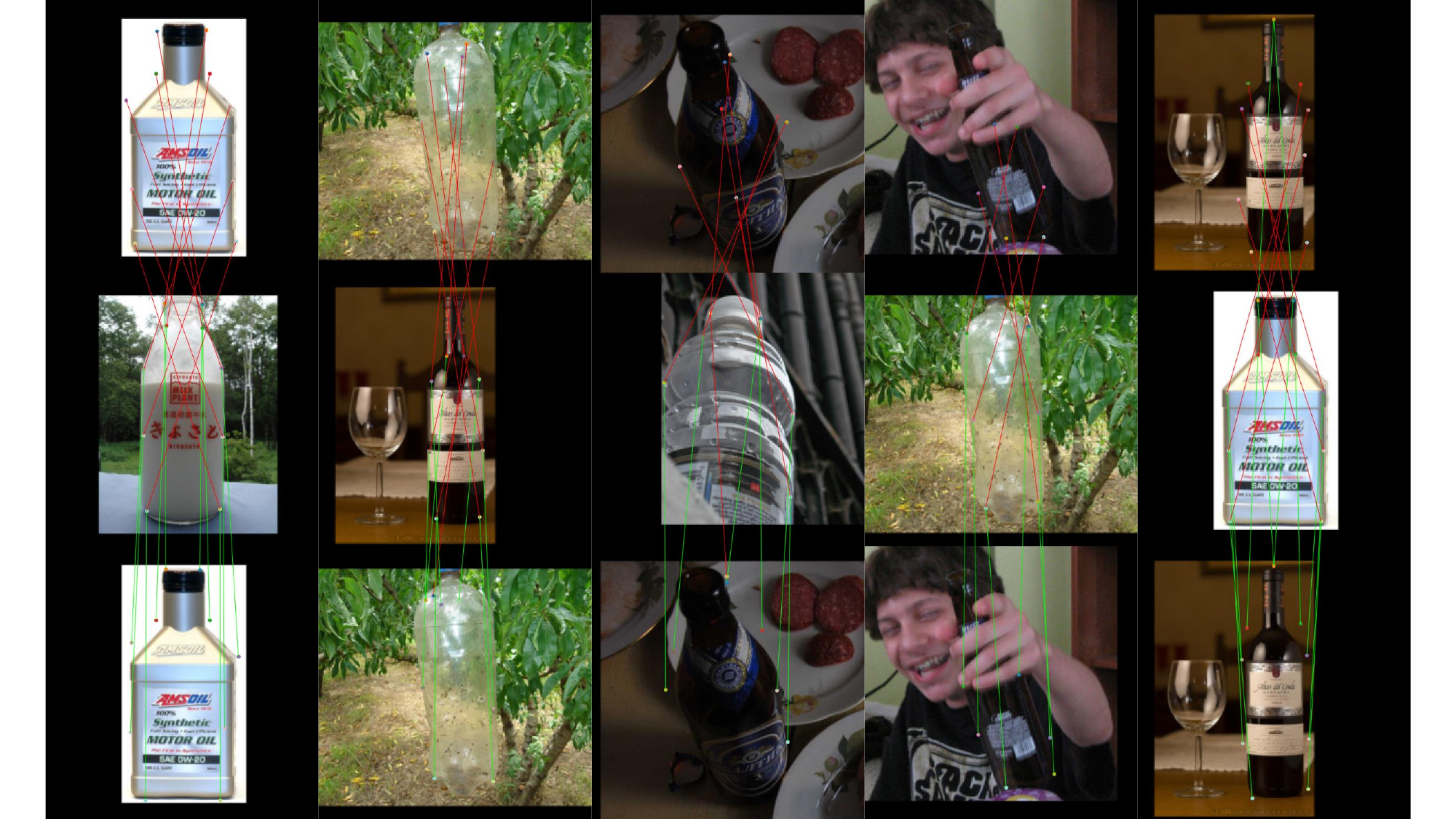}
    \includegraphics[width=0.49\linewidth]{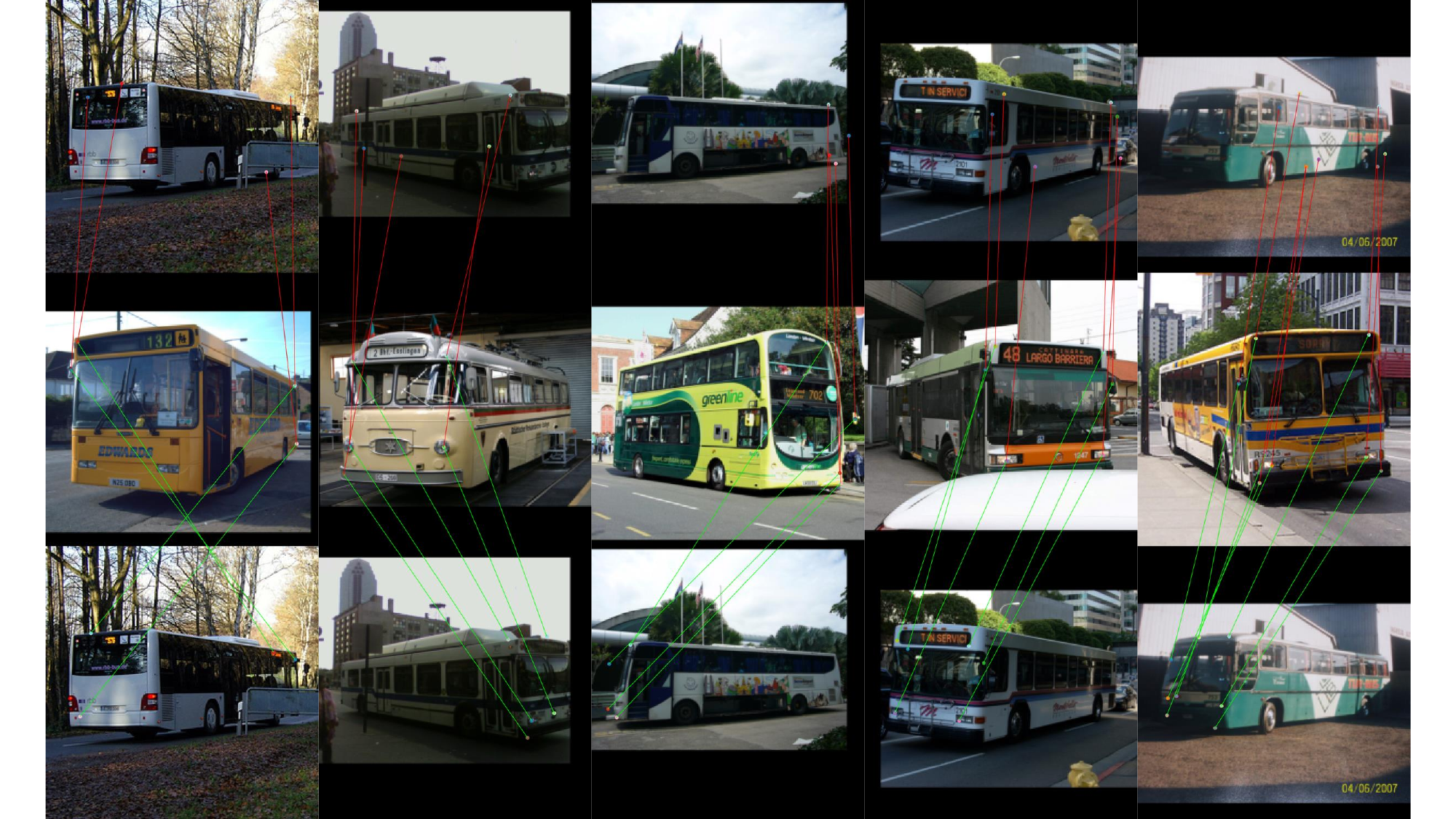}
    \hspace{0.001\linewidth}
    \includegraphics[width=0.49\linewidth]{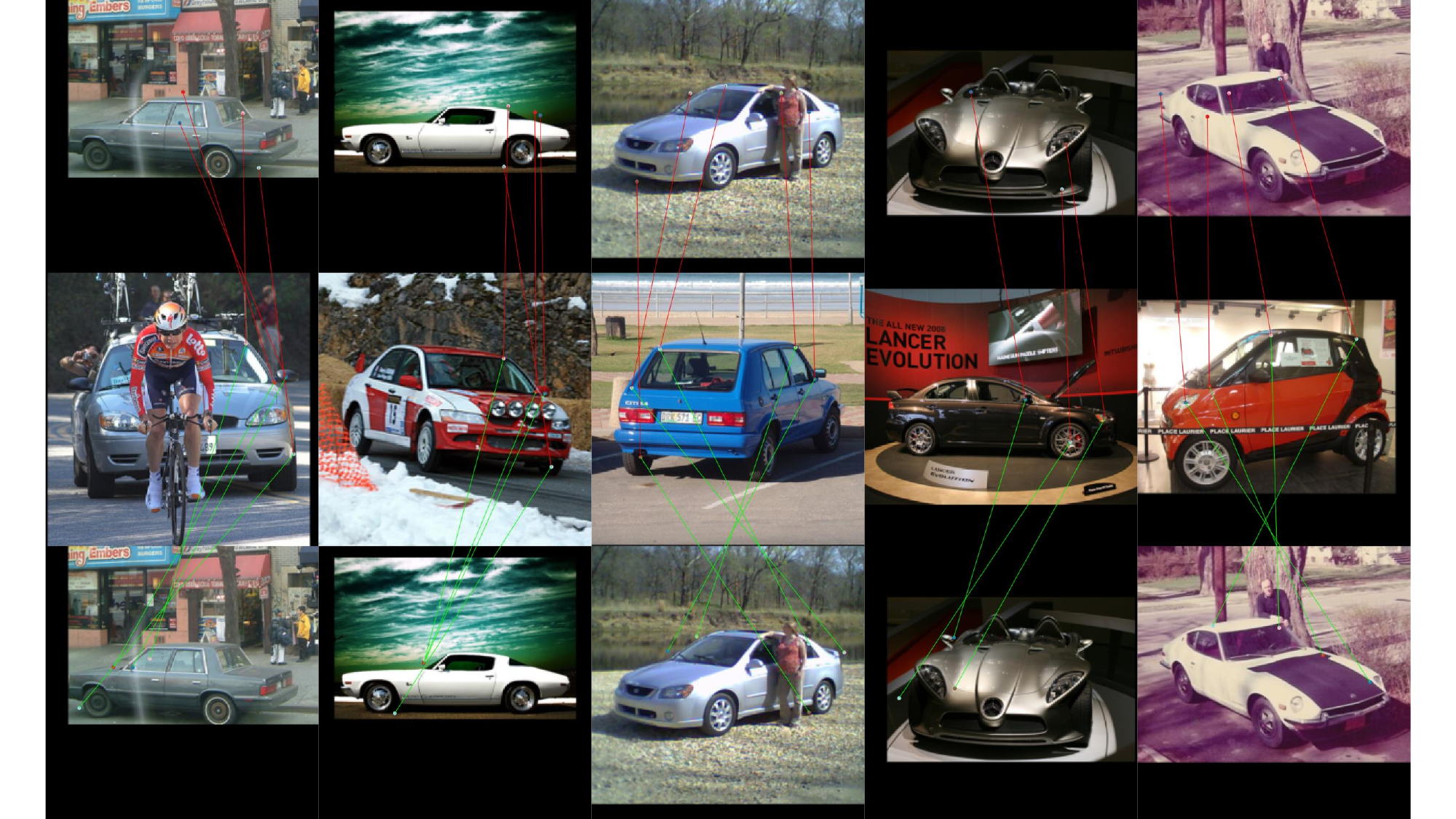}
    \includegraphics[width=0.49\linewidth]{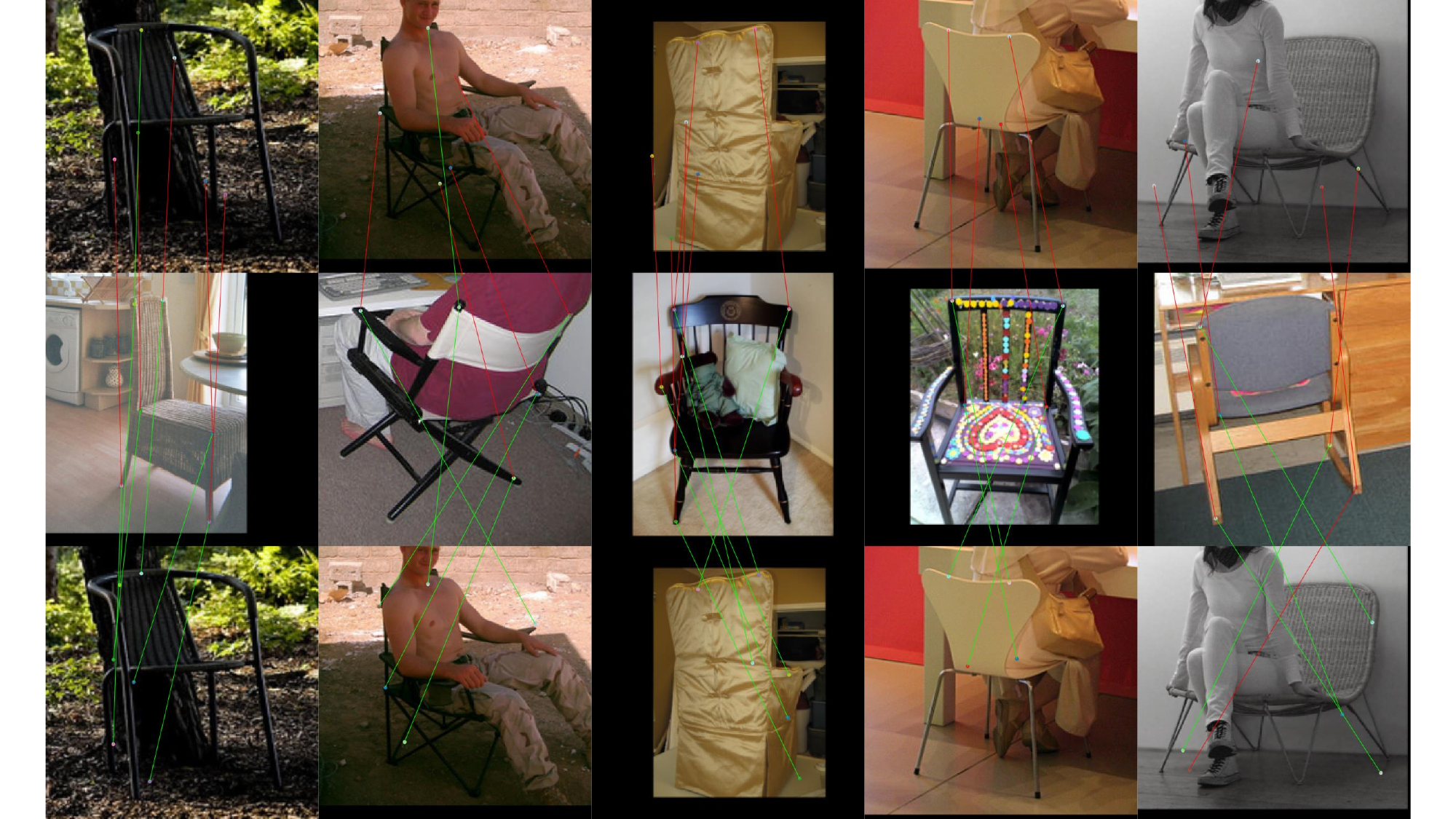}
    \hspace{0.001\linewidth}
    \includegraphics[width=0.49\linewidth]{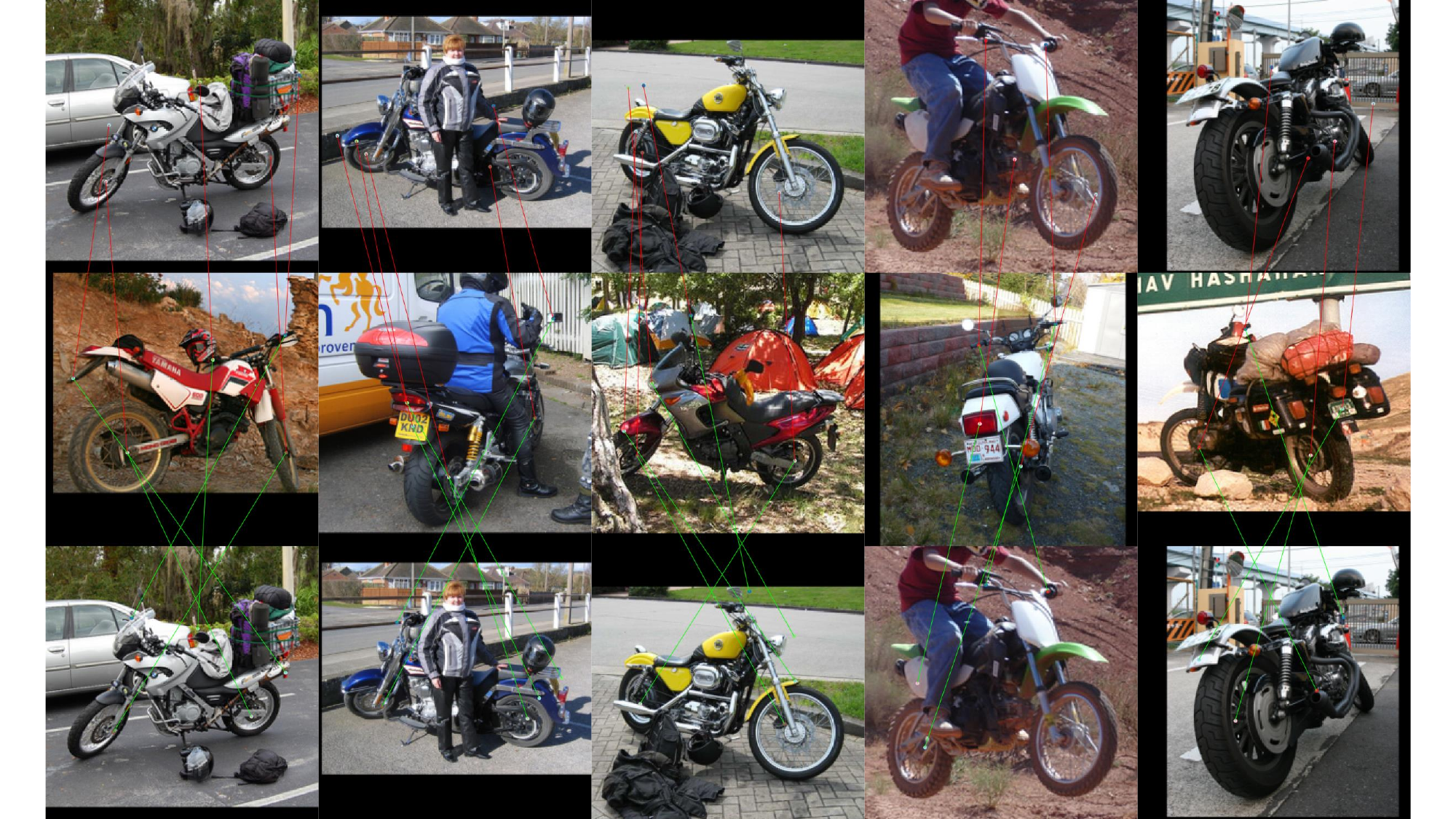}
    \includegraphics[width=0.49\linewidth]{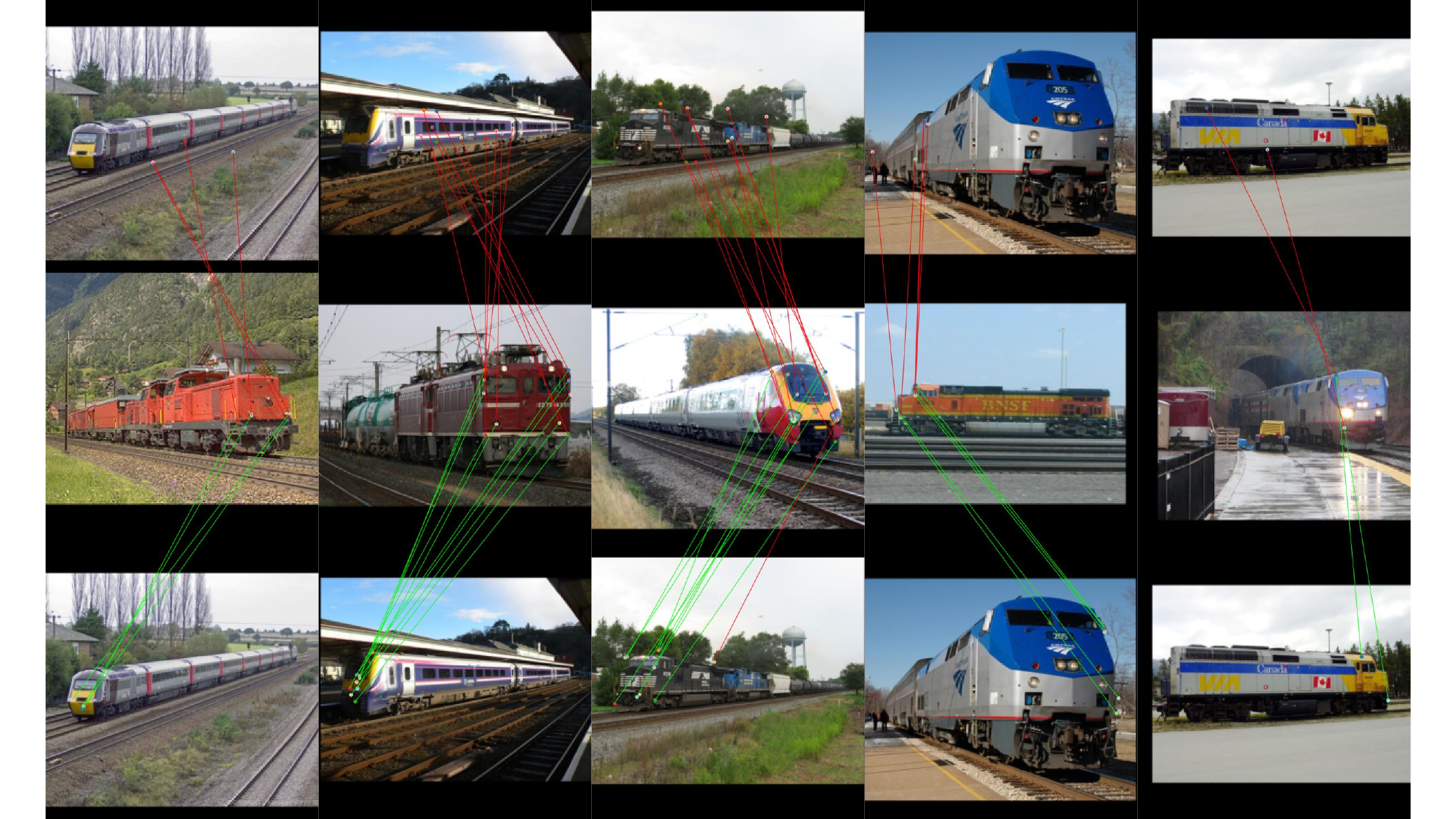}
    \hspace{0.001\linewidth}
    \includegraphics[width=0.49\linewidth]{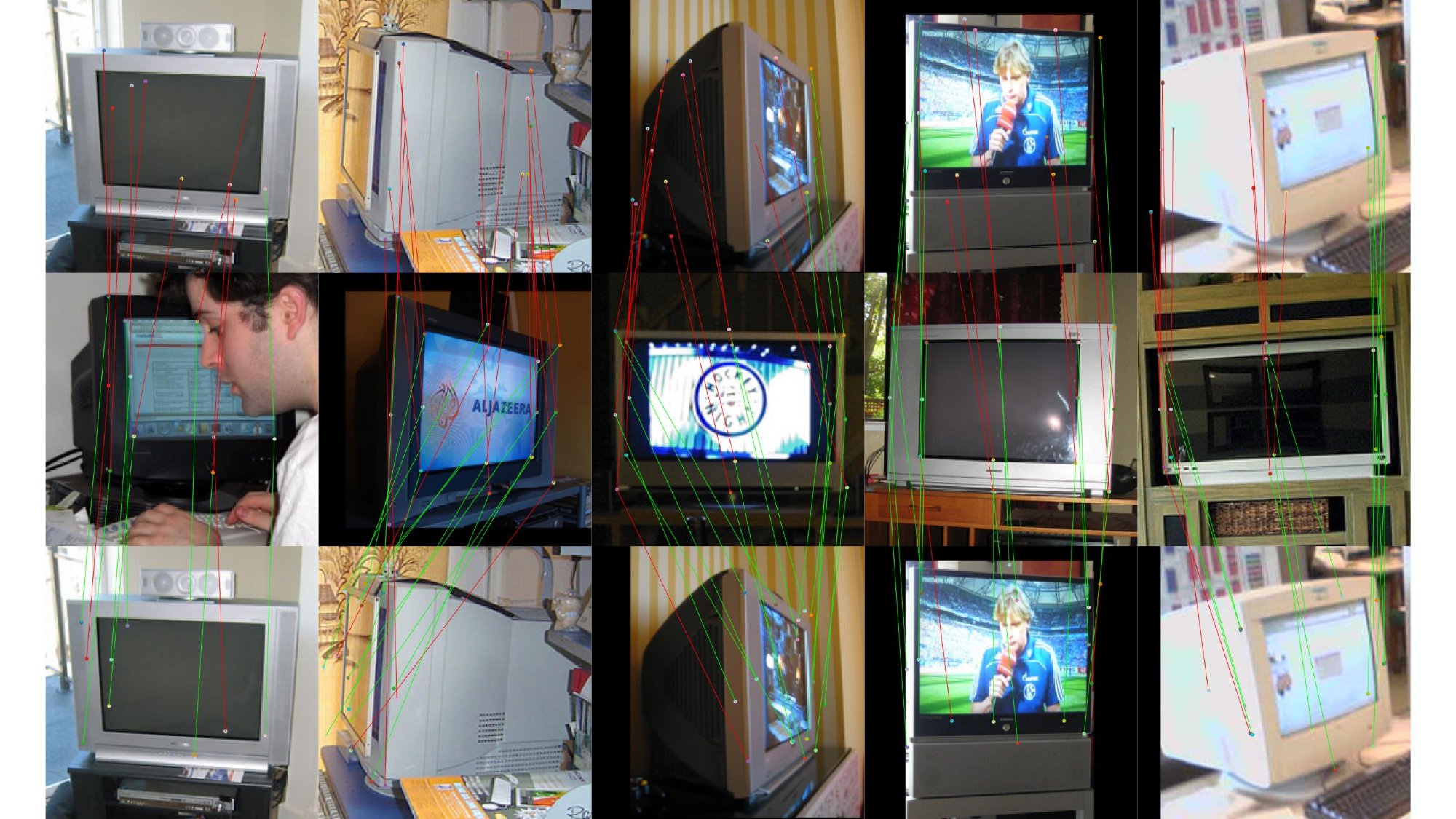}
    \caption{\textbf{Qualitative Examples on SPair-71k.} The first row shows the results of MagicPony, while the third row shows the results of our method. Green keypoint correspondence lines indicate correct predictions, whereas red lines indicate incorrect predictions. Our model improves semantic correspondence by reducing vertex drift and resolving ambiguities caused by symmetric structures and similar object parts.}
    \label{fig:qualitative_spair}
\end{figure}

In Figure \ref{fig:qualitative_spair}, we present comprehensive qualitative results on the SPair-71k dataset. The first row shows the results of MagicPony, while the third row shows the results of our method. Green keypoint correspondence lines indicate correct predictions, whereas red lines indicate incorrect predictions. Our model improves semantic correspondence by reducing vertex drift and resolving ambiguities caused by symmetric structures and similar object parts.

\end{document}